\documentclass[sigconf,nonacm]{aamas}

\usepackage{tikz}
\usetikzlibrary{arrows.meta,positioning}
\usepackage{xurl}
\usepackage{algorithm}
\usepackage{algpseudocode}
\usepackage{array}

\setcopyright{none}
\acmConference{}{}{}{}
\acmDOI{}
\acmISBN{}
\acmSubmissionID{}
\submissionType{}

\AtBeginDocument{%
  \fancypagestyle{standardpagestyle}{%
    \fancyhf{}
    \fancyhead[L]{\small Preprint}
    \fancyhead[R]{}
    \fancyfoot[C]{\thepage}

  }
  \fancypagestyle{firstpagestyle}{%
    \fancyhf{}
    \fancyhead[L]{\small Preprint}
    \fancyfoot[C]{\thepage}

  }
  \pagestyle{standardpagestyle}
}

\newcommand{\appendixArchitectureTable}{%
\begin{minipage}{\textwidth}
\centering
\captionof{table}{Architecture Of The Recurrent Agents.}
\label{tab:appendix-architecture}
\begin{tabular}{@{}p{0.22\textwidth}p{0.53\textwidth}p{0.19\textwidth}@{}}
\toprule
\textbf{Layer} & \textbf{Parameters} & \textbf{Activation} \\
\midrule
\multicolumn{3}{@{}l}{\textit{Input And Shared Recurrent Trunk}} \\
Vector Input (PQN) & Observation And Previous One-Hot Action; $d_{\mathrm{in}}=d_o+d_a$ & None \\
Vector Input (PPO) & Observation Only; $d_{\mathrm{in}}=d_o$ & None \\
Image Input (PQN) & CNN Features And Previous One-Hot Action; $d_{\mathrm{in}}=512+d_a$ & None \\
Input Projection & Linear, $d_{\mathrm{in}}\to256$ & None \\
\midrule
\multicolumn{3}{@{}l}{\textit{Recurrent Block}} \\
GRU & Input/Output Width $256\to256$ & Sigmoid \& Tanh \\
LRU & Input/Output Width $256\to256$ & None \\
Mixer Linear Layer & Linear, $256\to256$ & None \\
Mixer Normalization & LayerNorm Over $256$ Features, Without Affine Parameters & LeakyReLU \\
\midrule
\multicolumn{3}{@{}l}{\textit{Trunk Output And Readout}} \\
Output Projection & Linear, $256\to256$ & None \\
Readout Hidden Block & Linear, $256\to256$; LayerNorm With Affine Parameters; Repeat $K$ Times & LeakyReLU \\
Q Output (PQN) & Linear, $256\to d_a$; One Q-Value Per Action & None \\
Actor Output (PPO) & Linear, $256\to d_a$; Action Logits Or Gaussian Means & None \\
Critic Output (PPO) & Linear, $256\to1$; Scalar State Value & None \\
\bottomrule
\end{tabular}
\end{minipage}%
}

\newcommand{\appendixEncoderTable}{%
\begin{minipage}{\textwidth}
\centering
\captionof{table}{CNN Observation Encoder For POPGym Arcade.}
\label{tab:appendix-arcade-encoder}
\begin{tabular}{@{}p{0.22\textwidth}p{0.53\textwidth}p{0.19\textwidth}@{}}
\toprule
\textbf{Layer} & \textbf{Parameters} & \textbf{Activation} \\
\midrule
Input Scaling & $128\times128\times3$ RGB Image & None \\
Conv2D 1 & $3\to64$ Channels; $5\times5$ Kernel, Stride $2$; Output $64\times62\times62$ & LeakyReLU \\
MaxPool2D 1 & $2\times2$ Kernel, Stride $2$; Output $64\times31\times31$ & None \\
Conv2D 2 & $64\to128$ Channels; $3\times3$ Kernel, Stride $2$; Output $128\times15\times15$ & LeakyReLU \\
MaxPool2D 2 & $2\times2$ Kernel, Stride $2$; Output $128\times7\times7$ & None \\
Conv2D 3 & $128\to256$ Channels; $3\times3$ Kernel, Stride $2$; Output $256\times3\times3$ & LeakyReLU \\
MaxPool2D 3 & $3\times3$ Kernel, Stride $1$; Output $256\times1\times1$ & None \\
Conv2D 4 & $256\to512$ Channels; $1\times1$ Kernel, Stride $1$; Output $512\times1\times1$ & LeakyReLU \\
Flatten & $512\times1\times1\to512$ & None \\
\bottomrule
\end{tabular}
\end{minipage}%
}

\newcommand{\appendixMemoryChainPopgymTables}{%
\centering
\begin{minipage}[t]{0.48\textwidth}
\centering
\captionof{table}{PPO Hyperparameters For MemoryChain.}
\label{tab:appendix-training-memorychain}
\begin{tabular}{@{}p{0.64\columnwidth}>{\centering\arraybackslash}p{0.31\columnwidth}@{}}
\toprule
\textbf{Hyperparameter} & \textbf{Value} \\
\midrule
Initial Learning Rate & $3\times10^{-4}$ \\
Final Learning Rate & $10^{-10}$ \\
Discount $\gamma$ & $0.99$ \\
GAE $\lambda$ & $0.90$ \\
Gradient Clipping & $0.5$ \\
PPO Clip & $0.2$ \\
Entropy Coefficient & $0$ \\
Value-Loss Coefficient & $0.5$ \\
Number Of Environments & 64 \\
Rollout Length & 1024 \\
Minibatches Per Epoch & 16 \\
Update Epochs & 4 \\
Total Timesteps & $10$M \\
\bottomrule
\end{tabular}
\end{minipage}
\hfill
\begin{minipage}[t]{0.48\textwidth}
\centering
\captionof{table}{PQN Hyperparameters For POPGym.}
\label{tab:appendix-training-popgym}
\begin{tabular}{@{}p{0.64\columnwidth}>{\centering\arraybackslash}p{0.31\columnwidth}@{}}
\toprule
\textbf{Hyperparameter} & \textbf{Value} \\
\midrule
Initial LR (Trainable RNNs) & $10^{-5}$ \\
Initial LR (Fixed RNNs) & $5\times10^{-5}$ \\
Initial LR (MLP) & $5\times10^{-7}$ \\
Final Learning Rate & $10^{-10}$ \\
Discount $\gamma$ & $0.99$ \\
TD($\lambda$) & $0.95$ \\
Gradient Clipping & $10$ \\
Exploration $\epsilon$ & $1\to0.05$ \\
Number Of Environments & 16 \\
Rollout Length & 128 \\
Minibatches Per Epoch & 16 \\
Update Epochs & 4 \\
Total Timesteps & $20$M \\
\bottomrule
\end{tabular}
\end{minipage}%
}

\newcommand{\appendixArcadePlaygroundTables}{%
\centering
\begin{minipage}[t]{0.48\textwidth}
\centering
\captionof{table}{PQN Hyperparameters For POPGym Arcade.}
\label{tab:appendix-training-arcade}
\begin{tabular}{@{}p{0.64\columnwidth}>{\centering\arraybackslash}p{0.31\columnwidth}@{}}
\toprule
\textbf{Hyperparameter} & \textbf{Value} \\
\midrule
Initial LR (Trainable RNNs) & $5\times10^{-5}$ \\
Initial LR (Frozen RNNs) & $3\times10^{-4}$ \\
Final Learning Rate & $5\times10^{-7}$ \\
Discount $\gamma$ & $0.99$ \\
TD($\lambda$) & $0.65$ \\
Gradient Clipping & $10$ \\
Exploration $\epsilon$ & $1\to0.05$ \\
Number Of Environments & 16 \\
Rollout Length & 128 \\
Minibatches Per Epoch & 4 \\
Update Epochs & 4 \\
Total Timesteps & $20$M \\
\bottomrule
\end{tabular}
\end{minipage}
\hfill
\begin{minipage}[t]{0.48\textwidth}
\centering
\captionof{table}{PPO Hyperparameters For MuJoCo Playground.}
\label{tab:appendix-training-playground}
\begin{tabular}{@{}p{0.64\columnwidth}>{\centering\arraybackslash}p{0.31\columnwidth}@{}}
\toprule
\textbf{Hyperparameter} & \textbf{Value} \\
\midrule
Initial Learning Rate (Range) & $[1,5]\times10^{-5}$ \\
Final Learning Rate & $10^{-8}$ \\
Discount $\gamma$ & $0.995$ \\
GAE $\lambda$ & $0.95$ \\
Gradient Clipping & None \\
PPO Clip & $0.3$ \\
Entropy Coefficient & $0.01$ \\
Value-Loss Coefficient & $0.5$ \\
Number Of Environments & 16 \\
Rollout Length & 128 \\
Minibatches Per Epoch & 4 \\
Update Epochs & 4 \\
Total Timesteps & $60$M \\
\bottomrule
\end{tabular}
\end{minipage}%
}

\title[Does Learning Recurrent Networks Matter?]{Does Learning Recurrent Networks Matter in Partially Observable RL?}
\author{Zekang Wang}
\affiliation{
    \institution{University of Macau}
    \city{Macau SAR}\country{China}
}
\email{yc67543@um.edu.mo}

\author{Zhe He}
\affiliation{
    \institution{University of Macau}
    \city{Macau SAR}\country{China}
}
\email{yc67202@um.edu.mo}

\author{Gaojie Jin}
\authornote{Corresponding author.}
\affiliation{
    \institution{University of Macau}
    \city{Macau SAR}\country{China}
}
\email{gaojiejin@um.edu.mo}
\begin{abstract}
Reinforcement learning (RL) under partial observability typically relies on recurrent networks to encode temporal information, with the recurrent module trained jointly with the policy or value function.
This raises a natural question: is learning the recurrent network actually necessary? 
In this paper, we investigate whether a frozen recurrent network can provide sufficient memory for solving partially observable Markov decision processes (POMDPs).
Motivated by the theoretical foundations of Reservoir Computing (RC) and Generalized Reservoir Computing (GRC), we study a simple architecture that combines a fixed recurrent network with a trainable nonlinear output module.
We evaluate this approach across tasks with varying memory demands, observation modalities, and action spaces. 
Surprisingly, frozen recurrent networks support recall over hundreds of steps and achieve competitive control performance across diverse POMDPs, while using less than half the number of trainable parameters required by their fully trainable counterparts.
These results suggest that learning the recurrent memory module is not always necessary for solving POMDPs. 
Frozen recurrent networks therefore provide a simple, parameter-efficient, and surprisingly strong baseline for partially observable RL.
Our source code is available at \url{https://github.com/Aphlios/FRN-POMDP}.

\end{abstract}


\begin{document}
\maketitle

\begin{figure*}[!t]
\centering
\resizebox{\textwidth}{!}{
\begin{tikzpicture}[x=1cm,y=1.25cm,
  every node/.style={font=\sffamily\normalsize},
  block/.style={draw=black!45,line width=.65pt,rounded corners=2pt,
    align=center,minimum height=1.10cm,text width=2.6cm,inner sep=5pt,fill=white},
  fixed/.style={block,fill=black!9},
  learned/.style={block,draw=blue!55!black,fill=blue!7},
  flow/.style={-{Triangle[length=2mm,width=1.3mm]},line width=.85pt,
    draw=black!70,shorten >=1pt},
  update/.style={-{Triangle[length=2mm,width=1.3mm]},dashed,
    line width=1pt,draw=blue!65!black,shorten >=1pt},
  note/.style={font=\sffamily\normalsize\bfseries\boldmath,
    text=black!85,inner sep=2pt},
  update note/.style={note,text=blue!65!black},
  legend text/.style={anchor=west,font=\sffamily\normalsize,inner sep=0pt},
  layer/.style={block,minimum height=.52cm,text width=2.15cm,inner sep=3pt},
  learned layer/.style={layer,draw=blue!55!black,fill=blue!7},
  layer flow/.style={-{Triangle[length=1.4mm,width=1mm]},line width=.8pt,
    draw=black!70,shorten <=.5pt,shorten >=.5pt}]

\node[block,text width=1.75cm] (obs) at (0,0)
  {\textbf{Observation}\\$o_t$};
\node[learned,text width=2.4cm] (cnn) at (3.55,0)
  {\textbf{Trainable CNN}\\$E_{\psi}$\quad{\footnotesize(Optional)}};
\node[fixed,text width=3.6cm] (mem) at (7.85,0)
  {\textbf{Frozen Recurrent\\Network}\\$h_t=f_{\omega_0}(h_{t-1},u_t)$};
\node[learned] (read) at (12.25,0)
  {\textbf{MLP Readout}\\$Q_{\theta}(h_t,\cdot)$};
\node[block,text width=2.1cm] (env) at (16.1,0)
  {\textbf{Environment}};

\draw[flow] (obs)--node[note,below] {Images}(cnn);
\draw[flow] (cnn)--node[note,below] {$u_t$}(mem);
\draw[flow] (mem)--node[note,above] {$h_t$}(read);
\draw[flow] (read)--node[note,above] {$a_t$}(env);

\draw[flow,rounded corners=2pt] (1.55,0) -- (1.55,.98)
  -- node[note,above] {Vectors} (5.52,.98) -- (5.52,0);
\draw[flow,rounded corners=2pt] (env.north) -- (16.1,1.68)
  -- node[note,above] {Observation Feedback} (0,1.68) -- (obs.north);

\node[block] (loss) at (12.25,-2.4)
  {\textbf{PQN Learning}\\$\mathcal L_{\mathrm{TD}}$};
\draw[flow] ([xshift=-5.5mm]read.south)
  -- node[note,left] {Predictions} ([xshift=-5.5mm]loss.north);
\draw[flow,rounded corners=2pt] (env.south) -- (16.1,-2.4)
  -- node[note,above] {Rewards} (loss.east);
\draw[update] ([xshift=5.5mm]loss.north)
  -- node[update note,right] {Update $\theta$} ([xshift=5.5mm]read.south);
\draw[update,rounded corners=2pt] (loss.west) -- (9.75,-2.4)
  -- (9.75,-1.18) -- node[update note,above] {Update $\psi$ (Images)}
  (3.55,-1.18) -- (cnn.south);

\begin{scope}[shift={(0,-1.57)}]
\draw[draw=blue!55!black,fill=blue!7,line width=.65pt,rounded corners=1pt]
  (0,-.12) rectangle (.43,.12);
\node[legend text] at (.60,0) {Trainable Parameters};
\draw[draw=black!45,fill=black!9,line width=.65pt,rounded corners=1pt]
  (0,-.60) rectangle (.43,-.36);
\node[legend text] at (.60,-.48) {Fixed Parameters};
\draw[draw=black!45,fill=white,line width=.65pt,rounded corners=1pt]
  (0,-1.08) rectangle (.43,-.84);
\node[legend text] at (.60,-.96) {Parameter-free Elements};
\draw[flow] (5.30,-.24)--(5.96,-.24);
\node[legend text] at (6.20,-.24) {Data Flow};
\draw[update] (5.30,-.72)--(5.96,-.72);
\node[legend text] at (6.20,-.72) {Parameter Update};
\end{scope}

\draw[black!25,line width=.55pt] (18.05,2.30) -- (18.05,-3.40);
\node[font=\sffamily\normalsize\bfseries] at (19.9,2.10) {MLP Readout};
\node[inner sep=1pt] (readinput) at (19.9,1.47) {$h_t$};
\node[learned layer] (hiddenlinear) at (19.9,.66) {\textbf{Linear Layer}};
\node[learned layer] (hiddennorm) at (19.9,-.26) {\textbf{LayerNorm}};
\node[layer] (hiddenact) at (19.9,-1.18) {\textbf{LeakyReLU}};
\draw[layer flow] (readinput.south) -- (hiddenlinear.north);
\draw[layer flow] (hiddenlinear.south) -- (hiddennorm.north);
\draw[layer flow] (hiddennorm.south) -- (hiddenact.north);
\draw[black!55,line width=.65pt]
  (21.22,.95) -- (21.42,.95) -- (21.42,-1.47) -- (21.22,-1.47);
\node[anchor=west,inner sep=2pt] at (21.44,-.26) {$\times L$};
\node[learned layer,minimum height=.68cm] (outputlinear) at (19.9,-2.23)
  {\textbf{Linear Layer}\\Output};
\node[inner sep=1pt] (readoutput) at (19.9,-3.15) {$Q_\theta(h_t,\cdot)$};
\draw[layer flow] (hiddenact.south) -- (outputlinear.north);
\draw[layer flow] (outputlinear.south) -- (readoutput.north);
\end{tikzpicture}}
\caption{Overview of our PQN agent with frozen recurrence (left) and its nonlinear MLP Readout (right). We encode vector and visual observations differently. For visual control tasks, we train a Convolutional Neural Network (CNN) to extract and compress high-dimensional features. We optimize all trainable parameters using the PQN objective. The right panel shows the architecture of our simple MLP in detail. We also illustrate the architecture of the frozen recurrent network in Fig.~\ref{fig:frozen-recurrent-layer}.
}
\label{fig:hero}
\Description{Overview of our PQN agent with frozen recurrence (left) and its nonlinear MLP Readout (right). We encode vector and visual observations differently. For visual control tasks, we train a Convolutional Neural Network (CNN) to extract and compress high-dimensional features. We optimize all trainable parameters using the PQN objective. The right panel shows the architecture of our simple MLP in detail. We also illustrate the architecture of the frozen recurrent network in Fig.~\ref{fig:frozen-recurrent-layer}.}
\end{figure*}
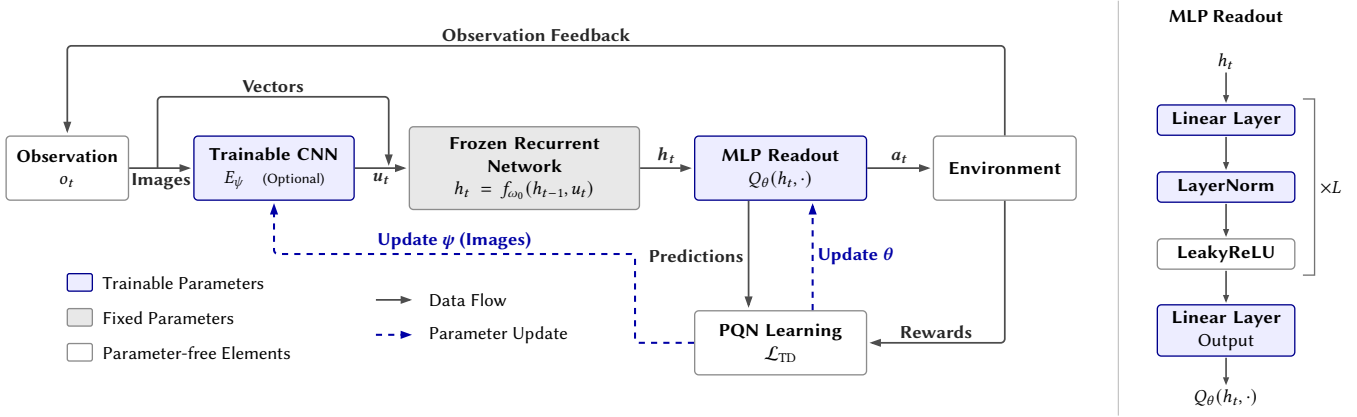

\section{Introduction}
\label{sec:introduction}
Reinforcement learning (RL) is commonly formulated as
a Markov Decision Process (MDP), in which the current state
contains all information needed for optimal decision making~\citep{sutton1998reinforcement}.
In many real-world settings, however, agents cannot directly observe the underlying state and instead receive partial or noisy observations that do not satisfy the Markov property.
Such problems are naturally modeled as partially observable Markov decision processes (POMDPs).
By incorporating memory, an agent can summarize its interaction history into a latent state representation that captures information missing from the current observation, thereby approximating a Markovian representation and enabling standard RL methods to be applied more effectively in partially observable settings~\citep{kaelbling1998planning}.

Recurrent neural networks (RNNs)~\citep{hochreiter1997long,elman1990finding,cho2014learning}, Transformers~\citep{vaswani2017attention,katharopoulos2020transformers,parisotto2020stabilizing,banino2022coberl,ni2023when,pramanik2024agalite}, and state-space models (SSMs)~\citep{gu2022efficiently,gu2024mamba,orvieto2023resurrecting,lu2023structured,lu2024rethinking,morad2024memoroids} have all been used as memory modules in model-free RL to summarize trajectory information and address POMDPs~\citep{ni2022recurrent}.
These memory models are usually trained jointly with the policy or value function. 
However, this raises the question that is learning the memory module actually necessary?
Several observations suggest that it may not always be. 
\citet{tallec2018reproducing} report that using an untrained Mixture-Density Recurrent Neural Network (MDRNN) from~\citet{NEURIPS2018_2de5d166} leads to only a 1.16\% performance increase in a model-based RL setting. More recently, \citet{gu2022parameterization} show that freezing the recurrent and input matrices of S4 and S4D can preserve performance on certain tasks. These findings are closely related to Reservoir Computing (RC)~\citep{jaeger2001echo,maass2002real,lukovsevivcius2009reservoir}, where a randomly initialized recurrent network serves as a fixed temporal feature extractor and only the output mapping is learned.
Despite these promising observations, it remains unclear whether a frozen recurrent network can serve as effective memory for general POMDPs. 
We therefore ask: can an agent solve POMDPs using a randomly initialized recurrent network as memory while learning only the policy or value function, and what design choices make this possible? 
As shown in Tab.~\ref{tab:fir-parameter-counts}, freezing the recurrent memory can substantially reduce the number of trainable parameters, motivating a simpler and more parameter-efficient approach to partially observable RL.

\begin{table}[!t]
\centering
\caption{Parameter counts for two types of recurrent agents we used. All hidden layers have the same width. Reduction is relative to the fully trainable agent.}
\label{tab:fir-parameter-counts}
\setlength{\tabcolsep}{3pt}
\begin{tabular*}{\columnwidth}{@{\extracolsep{\fill}}lrrrr@{}}
\toprule
Model & Total & Trainable & Frozen & Reduction \\
\midrule
\multicolumn{5}{l}{\textbf{PQN}} \\
\addlinespace[2pt]
GRU & 1,119,746 & 1,119,746 & 0 & --- \\
Frozen GRU & 1,119,746 & \textbf{331,778} & 787,968 & 70.37\% \\
\addlinespace[2pt]
LRU & 858,114 & 858,114 & 0 & --- \\
Frozen LRU & 858,114 & \textbf{331,778} & 526,336 & 61.34\% \\
\addlinespace
\multicolumn{5}{l}{\textbf{PPO}} \\
\addlinespace[2pt]
GRU & 1,385,221 & 1,385,221 & 0 & --- \\
Frozen GRU & 1,385,221 & \textbf{597,253} & 787,968 & 56.88\% \\
\addlinespace[2pt]
LRU & 1,123,589 & 1,123,589 & 0 & --- \\
Frozen LRU & 1,123,589 & \textbf{597,253} & 526,336 & 46.84\% \\
\bottomrule
\end{tabular*}
\Description{Parameter counts for two type of recurrent agents we used. All hidden layers have same width. Reduction is relative to the fully trainable agent.}
\end{table}

To answer this question, 
we need to revisit Reservoir Computing (RC) mentioned before, because it is the most mature paradigm to design and leverage frozen recurrent networks. 
RC typically consists of two components, a reservoir and a readout. Generally, reservoir is a randomly initialized RNN and its weights remain frozen during training. The Echo State Network (ESN)~\citep{jaeger2001echo} is a widely used architecture of RC. A key gaurantee in ESN is the Echo State Property (ESP) which indicates that the hidden state should be independent of its initial state and only retain information about the input history.
However, recurrent networks used in RL are usually different from the architecture of ESN, and the ESP is not a necessary condition for recurrent networks design. Generalized reservoir computing (GRC)~\citep{kubota2024reservoir} offers a broader perspective of ESP, which verify that a suitable readout can restore input history even the hidden state retain dependence from initialization. This suggests that ESP is unnecessary for using frozen recurrent network as memory in recurrent agent.

Through extensive experiments, we find that, similar to~\citep{tallec2018reproducing}, frozen recurrent networks can match the performance of trainable counterparts across a wide range of partially observable tasks,
including long-horizon delayed-recall problems, discrete and continuous control, and pixel-based POMDPs. 
Performance degradation appears only in a small number of task-specific cases, where freezing the recurrent network leads to modest losses.

Here, we summarize the main contributions in this paper as follows:
\textbf{(1)}~
We investigate whether frozen recurrent networks can serve as effective memory modules for solving POMDPs.
\textbf{(2)}~
We propose a simple and effective agent architecture that uses a frozen recurrent network as memory, reducing the number of trainable parameters in RL by more than half.
\textbf{(3)}~
Across a diverse set of POMDPs, we provide empirical evidence that learning the recurrent memory module is not always necessary for strong performance in partially observable RL.
\textbf{(4)}~
We further show that agents with frozen recurrent memory can successfully learn visual encoders for high-dimensional partially observable control tasks.

\section{Related Work}
\label{sec:related-work}
\paragraph{Recurrent Memory in Partially Observable RL}
Recurrent agents learn history-dependent representations through LSTM-based architectures and online recurrent updates~\citep{kapturowski2019recurrent,elelimy2024realtime}.
GTrXL, CoBERL, HCAM, DTQN, AGaLiTe, and RATE use attention or recurrent memory to process past information~\citep{parisotto2020stabilizing,banino2022coberl,lampinen2021towards,esslinger2022deep,pramanik2024agalite,cherepanov2026recurrent}.
S5, FFM, and SHM provide structured recurrence and memory updates~\citep{lu2023structured,morad2023reinforcement,le2025stable}.
MERLIN and MRA combine recurrent networks with external memory to store past experiences and retrieve information for decision making~\citep{wayne2018unsupervised,fortunato2019generalization}.
These approaches still involve learning many parameters within their recurrent memory. We explore an alternative approach by adapting existing trainable recurrent architectures into frozen recurrent memories for solving POMDPs.

\paragraph{Reservoir Computing and RL}
Reservoir computing separates a fixed recurrent backbone from a trainable readout~\citep{jaeger2001echo,maass2002real}. In POMDPs, ESN-based methods exploit this separation to construct history-dependent state representations for decision making~\citep{szita2006reinforcement,yamashita2022reinforcement}.
With trainable actor--critic readouts, ESNs can achieve competitive performance with learned recurrent memories on several memory-intensive tasks~\citep{mckee2024reservoir}. 
Related work has also explored fixed exponential memory traces for value prediction and control~\citep{eberhard2025memory}. However, these studies primarily focus on reservoir-specific architectures rather than frozen versions of standard recurrent networks. We take a simpler approach: directly freeze the recurrent unit of conventional architectures and train only the downstream policy or value function. Despite this minimal modification, the resulting agents achieve strong performance across a diverse range of POMDPs.


\paragraph{RL with Frozen Recurrent Networks}
Conventional frozen recurrent networks can also supply memory.
Agents with randomly initialized frozen LSTMs can learn visual interval-timing tasks~\citep{deverett2019interval}.
In robotic control, gated feature extraction combines an untrained GRU's summary of preceding observations with the current observation for TD3~\citep{bohm2023feature}.
However, agents with frozen RNNs and GRUs failed on accumulating-towers, range-count, and exact-count tasks under the reported A2C setup~\citep{mochizuki2023characterizing}.
However, these studies evaluate a limited range of POMDP environments. They also do not examine the theoretical basis of this phenomenon. We draw on GRC to guide the design of agents with frozen recurrent networks. Our simple architecture enables these agents to solve complex POMDPs.


\section{Background}
\label{sec:pre}
\subsection{Reinforcement Learning (RL)}
\begingroup
A Markov Decision Process (MDP)~\citep{sutton1998reinforcement} is a tuple $\mathcal{M} = \langle \mathcal{S}, \mathcal{A}, P, R, \gamma \rangle$, with state space $\mathcal{S}$, action space $\mathcal{A}$, state transition function $P(s_{t+1} \\ |s_t, a_t)$, reward function $R(r_t|s_t, a_t)$ and discount factor $\gamma$. In RL, at each timestep $t$, the agent observes a state $s_t \in \mathcal{S}$ and selects an action $a_t$ with respect to its policy $a_t \sim \pi(\cdot|s_t)$, which leads the agent to a new state $s_{t+1} \sim P(\cdot|s_t, a_t)$ while emitting a reward $r_t \sim R(\cdot|s_t, a_t)$. The Markov property states that the next state $s_{t+1}$ depends only on the current state $s_t$ and action $a_t$, as follows:
\begin{equation*}
    P(s_{t+1}|s_t, a_t)=P(s_{t+1}|s_t, a_t, s_{t-1},a_{t-1}, ..., s_0, a_0).
\end{equation*}
The agent's objective is to find an optimal policy $\pi^\star$ that maximizes the expected discounted sum of returns
\begin{equation}
    J(\pi, \mathcal{M})=\mathbb{E}_{\pi, \mathcal{M}}\left[\sum_{t=0}^\infty \gamma^t r_t\right].
\label{eq:rl-objective}
\end{equation}
\endgroup

\subsection{Partially Observable RL and Recurrent Memory}
A Partially Observable Markov Decision Process (POMDP) extends an MDP and is defined by the tuple $\mathcal{P}=\langle\mathcal{S},\mathcal{A},P,R,\Omega,O,\gamma\rangle$,
where $\mathcal{S}$, $\mathcal{A}$, $P$, $R$, and $\gamma$ follow the
definitions above, $\Omega$ is the observation space, and
$O(o_t|s_t)$ is the observation function. Unlike in an MDP, the
agent cannot directly observe the latent state $s_t$ and instead receives an
observation $o_t\sim O(\cdot|s_t)$. In that case, a single observation $o_t$ can not satisfy the Markov property, a trajectory $\mathbf{x}_t = (o_0, a_0, o_1, a_1, o_2, a_2, ..., a_{t-1}, o_t)$ is used instead~\citep{sutton1998reinforcement}.

The objective for an agent with memory model $f$ in a POMDP is therefore:
\begin{equation}
    J(f, \pi, \mathcal{P})=\mathbb{E}_{f,\pi, \mathcal{P}}\left[\sum_{t=0}^\infty \gamma^tr_t\right].
\label{eq:pomdp-objective}
\end{equation}
A recurrent memory model implements $f$ through a compact hidden state $\mathbf h_t$, updated from the previous state and current input~\citep{hausknecht2015deep,ni2022recurrent}:
\begin{equation}
\mathbf h_t=f_{\omega}(\mathbf h_{t-1},o_t),
\qquad \mathbf y_t=R_{\theta}(\mathbf h_t).
\label{eq:agent}
\end{equation}
Here, $o_t$ is the current observation, and the readout $R_\theta$ maps the hidden state $\mathbf h_t$ to action values or policy parameters.

A Gated Recurrent Unit (GRU) uses input- and state-dependent gates to control memory updates and retention~\citep{cho2014learning},
\begin{equation}
\begin{aligned}
\widetilde{\mathbf h}_t&=\tanh\!\left(\mathbf W o_t
+\mathbf U(\mathbf r_t\odot\mathbf h_{t-1})+\mathbf b\right),\\
\mathbf h_t&=(1-\mathbf z_t)\odot\mathbf h_{t-1}
+\mathbf z_t\odot\widetilde{\mathbf h}_t.
\end{aligned}
\label{eq:gru-background}
\end{equation}
Here, $\mathbf W$ and $\mathbf U$ are the input and recurrent weight matrices, $\mathbf b$ is a bias vector, and $\widetilde{\mathbf h}_t$ is the candidate hidden state. The reset gate $\mathbf r_t$ controls past-state information in the candidate, while the update gate $\mathbf z_t$ controls how much of the candidate replaces the previous hidden state. Both gates apply a sigmoid to affine functions of $o_t$ and $\mathbf h_{t-1}$; $\odot$ denotes elementwise multiplication.

A Linear Recurrent Unit (LRU) instead accumulates inputs in a complex-valued state through a diagonal linear recurrence~\citep{orvieto2023resurrecting},
\begin{equation}
\mathbf h_t=\mathbf\Lambda\mathbf h_{t-1}+\mathbf B o_t,
\qquad \mathbf\Lambda=\operatorname{diag}(\lambda_1,\ldots,\lambda_N).
\label{eq:lru-background}
\end{equation}
The magnitudes of $\lambda_i$ determine decay and their phases determine rotation, shaping memory retention. Real-valued output projections and nonlinear mixing may surround this linear cell.

\subsection{Reservoir Computing and the Echo State Property}
\label{sec:rc}
Reservoir Computing (RC) uses a dynamical system with randomly initialized recurrent networks to construct temporal features from an input sequence~\citep{jaeger2001echo,verstraeten2007experimental}. Its hidden state evolves with the inputs, and a learned readout maps this state to predictions or control tasks. In a classical Echo State Network (ESN), the state update is
\begin{equation}
\begin{aligned}
\mathbf h_t={}&(1-\alpha)\mathbf h_{t-1}\\
&+\alpha\tanh\!\left(\mathbf W_{\mathrm{in}} o_t
+\mathbf W_{\mathrm{rec}}\mathbf h_{t-1}+\mathbf b\right),
\end{aligned}
\label{eq:esn-background}
\end{equation}
where the leak rate $\alpha\in(0,1]$ controls the mixture of the previous state and nonlinear update. The input weights $\mathbf W_{\mathrm{in}}$, recurrent weights $\mathbf W_{\mathrm{rec}}$, and bias $\mathbf b$ remain fixed~\citep{jaeger2004harnessing}. A linear readout takes the form $\mathbf y_t=\mathbf W_{\mathrm{out}}\mathbf h_t+\mathbf b_{\mathrm{out}}$, although the readout can also be nonlinear.

The Echo State Property (ESP) describes the asymptotic independence of the hidden state from its initialization under a common input sequence~\citep{jaeger2001echo,yildiz2012re}. In the asymptotic forgetting formulation, trajectories $\mathbf h_t$ and $\mathbf h'_t$ driven by any admissible common input sequence converge from any pair of admissible initial states:
\begin{equation}
\lim_{t\to\infty}\|\mathbf h_t-\mathbf h'_t\|=0.
\label{eq:esp-background}
\end{equation}
For the leaky ESN, applying the $1$-Lipschitz bound for $\tanh$ to two such trajectories yields
\begin{equation}
\|\mathbf h_t-\mathbf h'_t\|_2
\leq\kappa\|\mathbf h_{t-1}-\mathbf h'_{t-1}\|_2,
\quad
\kappa=(1-\alpha)+\alpha\|\mathbf W_{\mathrm{rec}}\|_2.
\label{eq:esn-contraction}
\end{equation}
Thus $\kappa<1$ is sufficient for contraction in the induced Euclidean operator norm. A recurrent spectral radius below one alone does not supply this norm bound for a nonlinear reservoir.

Forgetting an initial state is different from forgetting an input. Eq.~\ref{eq:esp-background} holds the driving sequence fixed; control requires distinguishing histories with different driving sequences. The distinction is explicit in a linear system with fixed $\mathbf A,\mathbf B$:
\begin{equation}
\mathbf h_t=\sum_{k=0}^{t}\mathbf A^{t-k}\mathbf B o_k.
\label{eq:temporal-basis}
\end{equation}
\begin{figure*}[!t]
\centering
\includegraphics[width=0.94\textwidth]{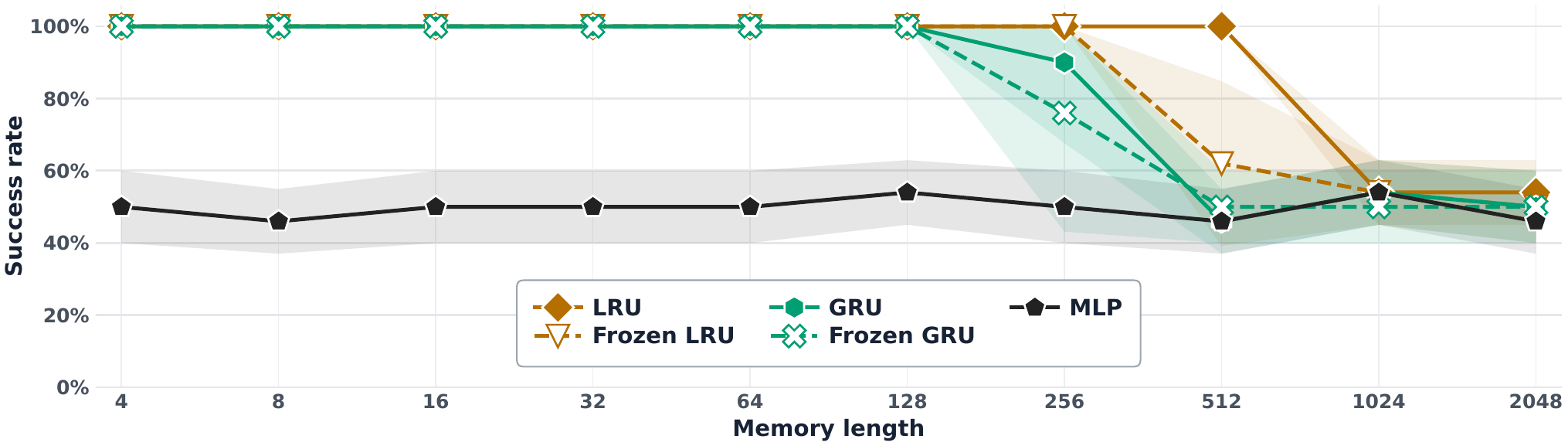}
\caption{Evaluation results on MemoryChain. The shaded regions report the range of success rates across 5 seeds and 10 evaluation episodesper seed.}
\label{fig:memorychain}
\Description{Evaluation results on MemoryChain. The shaded regions report the range of success rates across 5 seeds, and 10 evaluation per seed.}
\end{figure*}
\subsection{Generalized Reservoir Computing}
\label{sec:grc}
\label{sec:method}

Generalized Reservoir Computing (GRC) shifts the ESP requirement from the reservoir state to the output~\citep{kubota2024reservoir}. It seeks a time-invariant (TI) transformation that extracts a response determined by input history, even when the hidden state retains initialization dependence. For a readout $\mathbf y_t=R_\theta(\mathbf h_t)$, independence of the output from initialization requires
\begin{equation}
\lim_{t\to\infty}\|R_\theta(\mathbf h_t)-R_\theta(\mathbf h'_t)\|=0,
\label{eq:grc-output-forgetting}
\end{equation}
where $\mathbf h_t$ and $\mathbf h'_t$ receive identical inputs from different initial states. The resulting response has the form $\mathbf y_t^*=\Phi(o_t,o_{t-1},\ldots)$ after transients. TI means that the same mapping $\Phi$ applies at every time; the output can still vary with the input history.

A nonlinear readout can extract such a response by removing initialization-dependent components. For example, consider an oscillatory state $(X_t,Y_t)=(r_t\cos\varphi_t,r_t\sin\varphi_t)$ whose radius depends on input history while its phase retains initialization dependence. The transformation $X_t^2+Y_t^2=r_t^2$ removes the phase. GRC demonstrates related transformations using MLP readouts to recover past inputs from chaotic and physical reservoirs without ESP~\citep{kubota2024reservoir}.

\paragraph{Frozen LRU as Linear Reservoir}
Consider the LRU in Eq.~\ref{eq:lru-background}, with $\mathbf h_t\in\mathbb C^N$ and fixed matrices $\mathbf\Lambda$ and $\mathbf B$. Its stable diagonal parameterization~\citep{orvieto2023resurrecting} takes the form
\begin{equation}
\lambda_j=\exp\!\left(-\exp(\nu_j)+\mathrm i\varphi_j\right),
\qquad \rho=\max_j|\lambda_j|<1,
\label{eq:lru-stability-background}
\end{equation}
where $\nu_j$ and $\varphi_j$ are finite real parameters and $\mathrm i^2=-1$. Each eigenvalue has magnitude $\exp(-\exp(\nu_j))<1$. For two trajectories initialized at arbitrary finite states $\mathbf h_0$ and $\mathbf h'_0$ and driven by the same inputs, subtracting their updates cancels $\mathbf B o_t$ at every step. Writing $\Delta\mathbf h_t=\mathbf h_t-\mathbf h'_t$ gives
\begin{equation}
\begin{aligned}
\Delta\mathbf h_t&=\mathbf\Lambda\Delta\mathbf h_{t-1}
=\mathbf\Lambda^t\Delta\mathbf h_0,\\
\|\Delta\mathbf h_t\|_2&\leq\|\mathbf\Lambda^t\|_2\|\Delta\mathbf h_0\|_2
=\rho^t\|\Delta\mathbf h_0\|_2.
\end{aligned}
\label{eq:lru-esp-background}
\end{equation}
The equality $\|\mathbf\Lambda^t\|_2=\rho^t$ follows from diagonality. Since $\rho<1$, the initial-state difference vanishes as $t\to\infty$, establishing the ESP. For bounded inputs $\|o_t\|_2\leq M$ extending into the past, the corresponding history response is
\begin{equation}
\mathbf h_t^*=\sum_{k=0}^{\infty}\mathbf\Lambda^k\mathbf B o_{t-k},
\qquad \|\mathbf h_t^*\|_2\leq\frac{M\|\mathbf B\|_2}{1-\rho}.
\label{eq:lru-history-background}
\end{equation}
Thus, the frozen LRU cell is a stable linear reservoir satisfying the ESP. It is a conventional reservoir.

\paragraph{Frozen GRU as General Reservoir}
For a GRU, ESP depends on its parameters. Sufficient conditions for incremental stability ensure initial-state forgetting~\citep{bonassi2021stability}, but freezing the weights alone does not enforce these conditions. A two-dimensional construction illustrates how a frozen GRU can lack state ESP while admitting a readout with output ESP. Let $\mathbf h_t=(v_t,w_t)^\top\in[-1,1]^2$ and $o_t\in[-1,1]$. Set all gate weights and biases to zero, so $\mathbf z_t=\mathbf r_t=(1/2,1/2)^\top$. In Eq.~\ref{eq:gru-background}, choosing $\mathbf W=(1,0)^\top$, $\mathbf U=\operatorname{diag}(0,4)$, and $\mathbf b=0$ yields
\begin{equation}
\begin{aligned}
v_t&=\tfrac12v_{t-1}+\tfrac12\tanh(o_t),\\
w_t&=\tfrac12w_{t-1}+\tfrac12\tanh(2w_{t-1}).
\end{aligned}
\label{eq:gru-nonesp-background}
\end{equation}
The second coordinate has fixed points $0$ and $w_*\in(0,1)$, since $\tanh(2w)-w$ is positive just above zero and negative at one. Two trajectories with identical $v_0$ and inputs but with $w_0=0$ and $w'_0=w_*$ remain separated by $w_*$, violating state ESP. The first coordinate instead forgets its initialization. For updates starting at $t=1$, unrolling gives $v_t=2^{-t}v_0+\sum_{k=0}^{t-1}2^{-(k+1)}\tanh(o_{t-k})$. Extend the input sequence to $t\leq0$ with any bounded prehistory and write $o_{\leq t}=(o_t,o_{t-1},\ldots)$. Then
\begin{equation}
\begin{aligned}
\Phi(o_{\leq t})&=\sum_{k=0}^{\infty}2^{-(k+1)}\tanh(o_{t-k}),\\
|v_t-\Phi(o_{\leq t})|&\leq2^{1-t}\to0.
\end{aligned}
\label{eq:gru-readout-background}
\end{equation}
The bound follows from $|v_0|\leq1$ and the geometric tail of the series. The linear readout $y_t=v_t$ therefore approaches a nonconstant function of input history. For a nonlinear readout $y_t=\ell(v_t)$, 
where
$\ell(v)=\max(v,0.01v)$, the $1$-Lipschitz property of $\ell$ gives
\[
\bigl|y_t-\ell(\Phi(o_{\leq t}))\bigr|
\leq \bigl|v_t-\Phi(o_{\leq t})\bigr|
\longrightarrow 0.
\]
Since $\ell$ is strictly increasing, the function of
input history remains nonconstant. Both readouts discard the initialization-dependent coordinate, illustrating the GRC principle~\citep{kubota2024reservoir}.

\begin{figure*}[htpb]
\centering
\includegraphics[width=\textwidth]{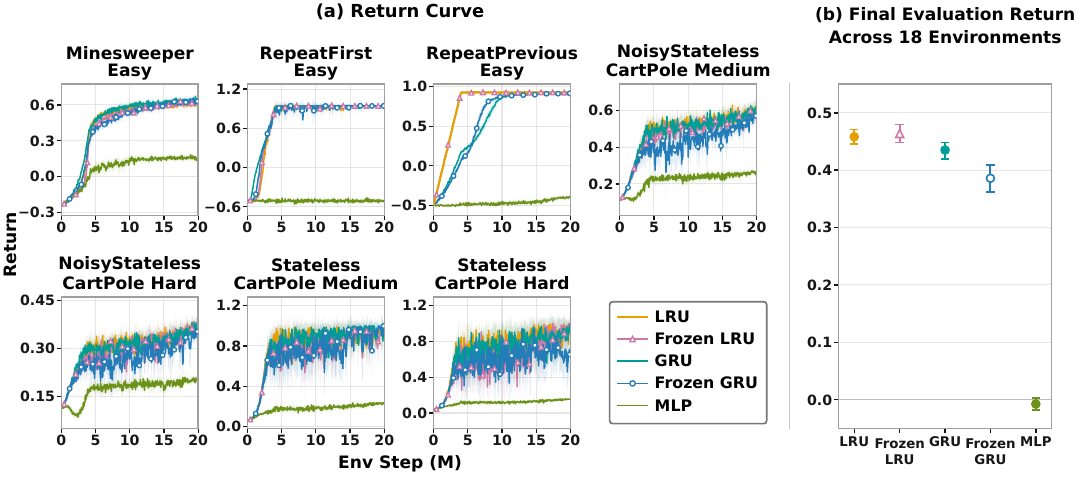}
\caption{POPGym Results: (a) learning curves on seven selected environments, with 10 seeds per environment; (b) final evaluation returns over all 18 environments, with 10 seeds per environment and 10 evaluation episodes per seed.}
\label{fig:return-curves}
\Description{POPGym Results: (a) learning curves on seven selected environments, with 10 seeds per environment; (b) final evaluation returns over all 18 environments, with 10 seeds per environment and 10 evaluation episodes per seed.}
\end{figure*}
\begin{algorithm}[htpb]
\caption{Constructing and Using Frozen Recurrent Agents}
\label{alg:fixed-recurrence}
\begin{algorithmic}[1]
\Ensure Agent and its trainable parameter set $\Theta_{\mathrm{learn}}$
\Statex $F_{\omega_0,\phi}$ is the recurrent memory network: $\omega_0$ contains all recurrent unit parameters at initialization; $\phi$ denotes the trainable mixer parameters.
\State Initialize $F_{\omega_0,\phi}$ and $R_\theta$
\State Use \textsc{FrozenStep} for each recurrent unit
\State $\Theta_{\mathrm{learn}}\gets\{\phi,\theta\}$
\State $E\gets\mathrm{Identity}$
\If{observations are images}
    \State Initialize CNN $E_\psi$; set $E\gets E_\psi$
    \State $\Theta_{\mathrm{learn}}\gets\Theta_{\mathrm{learn}}\cup\{\psi\}$
\EndIf
\State Use $E(o_t)$ as the input to $F_{\omega_0,\phi}$
\State Keep $\omega_0$ fixed; update only $\Theta_{\mathrm{learn}}$ with the RL objective
\State \Return agent $(E,F_{\omega_0,\phi},R_\theta)$ and $\Theta_{\mathrm{learn}}$
\Statex
\Function{FrozenStep}{$c,\omega_{c,0},\mathbf h_{t-1},\mathbf u_t,e_t$}
    \Statex \hspace{\algorithmicindent}$c$: a single recurrent unit (e.g., one GRU or LRU unit).
    \Statex \hspace{\algorithmicindent}$\omega_{c,0}$: unit $c$'s initialized parameter block in $\omega_0$.
    \Statex \hspace{\algorithmicindent}$t$: time step; $\mathbf h_{t-1}$: the cell's previous hidden state.
    \Statex \hspace{\algorithmicindent}$\mathbf u_t$: input features supplied to $c$ at step $t$.
    \Statex \hspace{\algorithmicindent}$e_t\in\{0,1\}$: episode start flag.
    \If{$e_t=1$}
        \State $\mathbf h_{t-1}\gets\Call{InitialState}{c}$ \Comment{unit's reset state}
    \EndIf
    \State $\mathrm{SG}\gets\texttt{jax.lax.stop\_gradient}$
    \State $\bar\omega_c\gets\texttt{jax.tree.map}(\mathrm{SG},\omega_{c,0})$
    \State $(\mathbf h_t,\mathbf x_t^{\mathrm{out}})\gets c_{\bar\omega_c}(\mathbf h_{t-1},\mathbf u_t)$
    \State \Return $\mathbf h_t,\mathbf x_t^{\mathrm{out}}$ \Comment{Updated state, output features}
\EndFunction
\end{algorithmic}
\end{algorithm}



\section{Experiments}
\label{sec:experiments}
Our experiments aim to answer the following research questions:

\begin{itemize}
    \item \textbf{RQ1:} Can frozen recurrent networks retain input over long horizons in RL? (Sec.~\ref{sec:memorychain})
    \item \textbf{RQ2:} Can frozen recurrent networks solve vector observation discrete POMDPs? (Sec.~\ref{sec:popgym})
    \item \textbf{RQ3:} Can frozen recurrent networks solve vector observation continuous POMDPs? (Sec.~\ref{sec:mujoco})
    \item \textbf{RQ4:} Can frozen recurrent networks solve high-dimensional image observed POMDPs? (Sec.~\ref{sec:arcade})
    \item \textbf{RQ5:} Which learning matter in frozen recurrent networks? (sec.~\ref{sec:learning})
\end{itemize}

\subsection{Experimental Setup}
\label{sec:protocol}

\paragraph{Baselines.} \citet{morad2023popgym} evaluated serveral memory archiectures and found that GRU is the most performant. Recently, \citet{wang2026investigating} and \citet{lu2024rethinking} found that LRU is the most powerful and efficient state-space memory architecture for RL. We therefore pick both of them as our baselines. For the frozen variants, we simply apply Algo.~\ref{alg:fixed-recurrence} to the recurrent agent. Here, we use JAX-style pesudo code as an example.

\paragraph{Agent Architecture.}
We show our recurrent PQN agent architecture in Fig.~\ref{fig:hero}. We use a simple MLP as readout, and each MLP consists of $L$ blocks with a Linear Layer, a Layer Normalization~\citep{ba2016layer}, and LeakyReLU activation. We only stop the gradient of parameters in the recurrent unit, instead of cutting off gradient computation of recurrent models, so our agent still can learn a visual encoder to compress high-dimensional image observations. We update all the trainable parameters with the PQN learning target. Fig.~\ref{fig:frozen-recurrent-layer} shows our detailed architecture of frozen recurrent networks. Each frozen recurrent unit is followed by a trainable feature mixer. For PPO, we only replace the Q-value readout MLP with same separate actor and critic MLP heads.

\begin{figure*}[t]
\centering
\includegraphics[width=\textwidth]{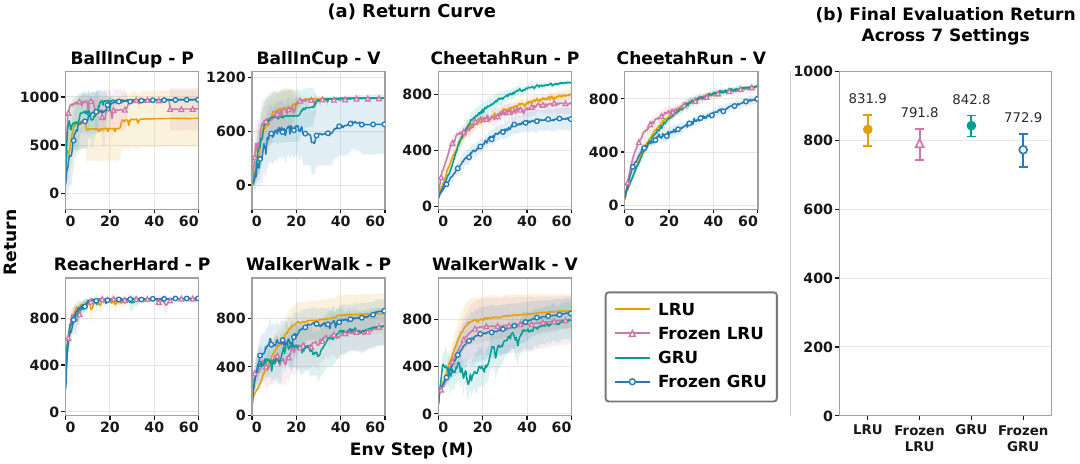}
\caption{MuJoCo Playground Results: (a) learning curves across seven \emph{position only} (P) and \emph{velocity only} (V) settings, with means and confidence intervals over 10 seeds; (b) final evaluation return averaged across the seven settings over 10 seeds. Each seed is evaluated on 10 episodes after training.}
\label{fig:playground-curves}
\Description{MuJoCo Playground Results: (a) learning curves across seven position only (P) and velocity-only (V) settings, with means and confidence intervals over 10 seeds; (b) final evaluation return averaged across the seven settings over 10 seeds. Each seed is evaluated on 10 episodes after training}
\end{figure*}

\paragraph{MemoryChain.}
MemoryChain is a classical one-bit delayed-recall task from bsuite~\citep{Osband2020Behaviour}. We use the implementation from Gymnax~\citep{gymnax2022github} for faster rollout. We train eight delay configurations with Proximal Policy Optimization (PPO)~\citep{schulman2017proximal}, using $10$M env steps and 5 seeds at each delay. Each seed receives ten evaluation episodes.
\paragraph{POPGym.}


Partially Observable Process Gym (POPGym) is a well-known benchmark for evaluating memory in partially observable RL~\citep{morad2023popgym}. We use POPJym~\citep{lu2023structured}, a JAX implementation of POPGym. Our evaluation includes six tasks: Minesweeper, NoisyStatelessCartPole, RepeatFirst, RepeatPrevious, and StatelessCartPole. Each task has three difficulty levels, giving 18 settings. We train each agent using PQN~\citep{gallici2025simplifying} with a budget of $20$ million environment steps with 10 seeds per run.
\paragraph{POMDP MuJoCo Playground.}
We also hope to apply our framework to robot control. We therefore evaluate it on the DeepMind Control Suite tasks reimplemented in MuJoCo Playground~\citep{zakka2025playground}. We use four environments: BallInCup, CheetahRun, ReacherHard, and WalkerWalk.  We introduce partial observability by modifying the original observations. The \emph{Positional Only} setting retains position information and hides velocity information. The \emph{Velocity Only} setting retains velocity information and hides position information.  Following the original paper, we use a budget of $60$ million environment steps per run. We train with PPO~\citep{schulman2017proximal} using a shared Actor-Critic trunk. Each configuration uses 10 seeds.

\paragraph{POPGym Arcade.}
We extend the evaluation to pixelated POMDPs with POPGym Arcade~\citep{wang2026investigating}. This benchmark provides both fully observable and partially observable Atari-style environments implemented in JAX. Our experiments cover the partially observable variants of BattleShip, CountRecall, MineSweeper, and Navigator. Agents receive $128\times128\times3$ RGB images as the observation from environments. To reduce computational cost and improve training efficiency, we train the models with PQN~\citep{gallici2025simplifying}. The training budget is $20$ million environment steps per seed, with five training seeds per configuration.

\begin{figure*}[t]
\centering
\includegraphics[width=\textwidth]{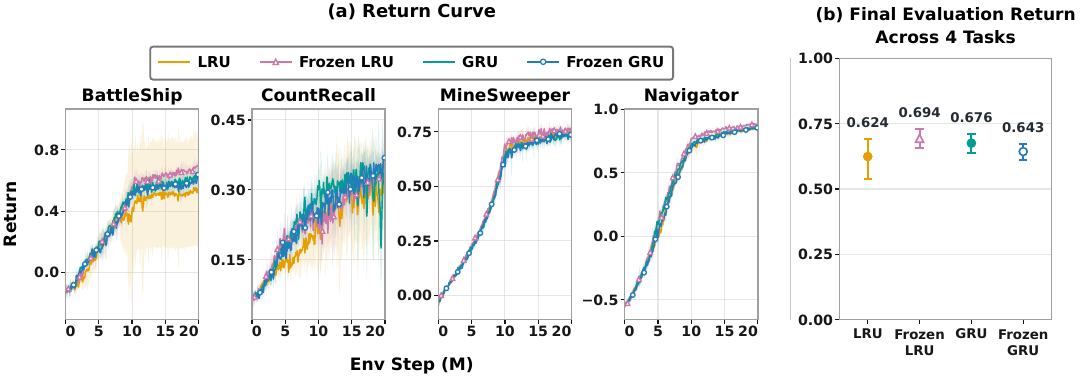}
\caption{POPGym Arcade Results: (a) learning curves on four partially observable tasks with means and confidence intervals over 5 seeds; (b) final evaluation return averaged across the four tasks over 5 seeds. Each seed is evaluated on ten episodes. }
\label{fig:arcade-curves}
\Description{POPGym Arcade Results: (a) learning curves on four partially observable tasks with means and confidence intervals over 5 seeds; (b) final evaluation return averaged across the four tasks over 5 seeds. Each seed is evaluated on ten episodes.}
\end{figure*}

\paragraph{Evaluation.}
We evaluate every agent after training. For each seed, we run 10 episodes and compute the mean return and variance to make the final evaluation return plots in Fig.~\ref{fig:memorychain}, \ref{fig:return-curves}, \ref{fig:playground-curves} and \ref{fig:arcade-curves}. 

\subsection{RQ1: Long-Horzion Recall
}
\label{sec:memorychain}
We use MemoryChain to compare the recall performance of Frozen and trainable recurrent networks. In this environment, the agent receives a single-bit cue at the first time step of the episode, and then receives a sequence of zero observations for a number of delay steps. The agent receives a positive reward of $+1$ if it outputs the correct initial cue at the last time step, and $-1$ otherwise. We sweep the delay $2^k$ for $k \in \{ 2, ..., 11\}$ and compare MLP, GRU, LRU, frozen GRU and frozen LRU. Here, we introduce MLP as a memoryless baseline. We implement recurrent PPO with JAX~\citep{jax2018github} and Equinox~\citep{kidger2021equinox} to accelerate the training process.

In Fig.~\ref{fig:memorychain}, we plot the success rate of evaluation results. LRU achieves 100\% success rate at delays up to 512 length, as expected for a state-of-the-art memory model in POMDPs. More surprisingly, frozen LRU gets 100\% success rate at delays up to 256 length close to LRU. Similarly, the recall ability of frozen GRU also matches that of GRU, both achieve 100\% success rate at delay of 128 length. Although the performance of frozen LRU and GRU does not exceed that of their trainable counterparts, these results still show the evidence that frozen recurrent networks have the ability to retain a cue over long horizons, indicating that simply freezing the parameters of the recurrent unit does not hurt too much of the memory retention of recurrent networks with RL objective.

\subsection{RQ2: Vector Observation Discrete POMDPs}
\label{sec:popgym}
POPGym introduces partial observability to some simple RL tasks, making them more challenging and long-term memory demanding. For example, in MineSweeper, the agent can only observe the number of mines nearby instead of the actual grid, which requires the agent to remember which grids have been revealed before. We use 18 environments with approximately 1000 runs on POPGym to test the performance of Frozen GRU and LRU compared with their trainable counterparts as well as MLP. We plot seven representative selected learning curves in Fig.~\ref{fig:return-curves}(a) and the final evaluation returns for all environments in Fig.~\ref{fig:return-curves}(b). We present only informative results and exclude very simple tasks that almost all recurrent networks can solve.

Frozen recurrent networks achieve competitive performance while training substantially fewer parameters. As shown in Tab.~\ref{tab:fir-parameter-counts}, Frozen LRU trains only approximately 40\% of the total agent parameters, yet achieves a final evaluation return of 0.464, compared with 0.458 for LRU (Fig.~\ref{fig:return-curves}(b)). This corresponds to a relative increase of 1.23\%. Frozen GRU trains only approximately 30\% of the total agent parameters and achieves a final evaluation return of 88.70\% for GRU. These results show that updating all recurrent parameters is unnecessary for effective control on the evaluated POMDP tasks. In Fig.~\ref{fig:return-curves}(a), the learning curves of Frozen LRU and LRU closely overlap on most tasks, indicating similar training progress and final performance. Although Frozen GRU often achieves lower returns than GRU during training, its final returns are comparable to GRU.
These results demonstrate that frozen recurrent networks can serve as effective memory models to solve discrete POMDPs and achieve competitive performance.

\subsection{RQ3: Vector Observation Continuous POMDPs
}
\label{sec:mujoco}
To further test the applicability of frozen recurrent networks in POMDPs, we also conducted experiments on partially observable continuous control tasks. We use MuJoCo Playground~\citep{zakka2025playground}, the latest JAX reimplementation of MuJoCo~\citep{todorov2012mujoco}. It consists of several challenging tasks in which agents need to control a multi-joint dynamical body. We introduce \emph{Position Only} and \emph{Velocity Only} settings widely used in prior work for evaluating RL agents with memory~\citep{ni2022recurrent,meng2021memory,ni2023when}, which are denoted as P and V in Fig.~\ref{fig:playground-curves}, respectively. We implement wrappers to mask the positional or velocity information in their observation.

Similarly, frozen recurrent networks also achieve competitive performance in partially observable continuous control tasks. In our PPO agents, Frozen LRU and Frozen GRU train approximately 53\% and 43\% of the total agent parameters, respectively (Tab.~\ref{tab:fir-parameter-counts}). Across the seven observation settings in Fig.~\ref{fig:playground-curves}(b), Frozen LRU achieves a final evaluation return of 791.8, compared with 831.9 for LRU, retaining 95\% of LRU's return. Frozen GRU achieves 772.9, compared with 842.8 for GRU, retaining 92\% of GRU's return.
The learning curves in Fig.~\ref{fig:playground-curves}(a) show similar returns for all four agents on ReacherHard. On WalkerWalk, Frozen GRU has higher mean training returns than GRU near the end of training under both \emph{Position Only} and \emph{Velocity Only} settings. The learning curves also show lower returns for Frozen LRU than LRU on WalkerWalk and for Frozen GRU than GRU on CheetahRun at the end of training. These results demonstrate that frozen recurrent networks can provide sufficient and efficient memory to solve such continuous POMDPs.

\subsection{RQ4: Partially Observable Visual Control}
\label{sec:arcade}
In real-world scenarios, we usually regard most of the tasks as partially observable visual control tasks, in which agents need to make optimal decisions with image observations. With respect to Fig.~\ref{fig:hero} and Sec.~\ref{sec:protocol}, we can use gradients through a frozen recurrent network to learn a CNN as observation encoder to solve visual control POMDPs.

We evaluate our agents on POPGym Arcade~\citep{wang2026investigating}, a JAX implemented extension of POPGym that provides various partially observable visual control tasks. It includes very challenging and high-dimensional visual POMDPs. For example, in the partially observable versions of BattleShip and MineSweeper, each observation reveals information only about the current grid cell. The agent must therefore extract features from these high-dimensional images, retain information about previously observed cells in memory, and retrieve it to guide decisions.

Notably, frozen recurrent networks also achieve competitive performance with learned visual representations. Across the four tasks in Fig.~\ref{fig:arcade-curves}(b), Frozen LRU achieves a final evaluation return of 0.694, compared with 0.624 for LRU, representing a relative increase of 11.2\% with lower variance. Frozen GRU achieves 0.643, retaining 95.2\% of GRU's final evaluation return of 0.676. In Fig.~\ref{fig:arcade-curves}(a), the learning curves of all four agents approach similar returns on MineSweeper and Navigator. Frozen LRU reaches higher mean training returns than LRU on BattleShip, while the training curves on CountRecall fluctuate more but remain comparable at the end. 
These results demonstrate that frozen recurrent networks
can work with learned CNN encoders to solve challenging and high-dimensional visual control POMDPs. Moreover, Frozen LRU can even outperform its trainable counterpart, making it a promising baseline for partially observable RL. 



\begin{figure}[!t]
\centering
\includegraphics[width=\columnwidth]{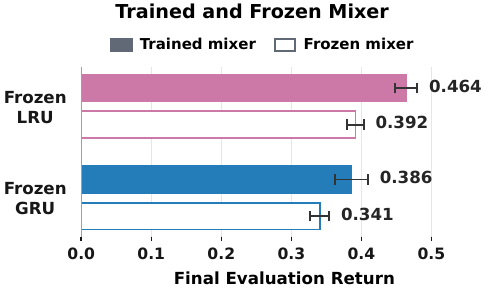}
\caption{Comparison with trainable or frozen mixer. Bars show final evaluation return across 18 POPGym tasks, with 10 seeds and 10 evaluation episodes per seed.}
\label{fig:mixer-training}
\Description{Comparison with trainable or frozen mixer. Bars show final evaluation return across 18 POPGym tasks, with 10 seeds and 10 evaluation episodes per seed.}
\end{figure}
\subsection{RQ5: Learning the Mixers Matter}
\label{sec:learning}

\begin{figure}[!t]
\centering
\includegraphics[width=\columnwidth]{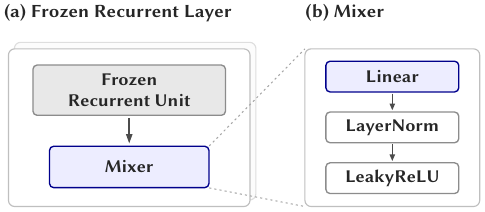}
\caption{Frozen recurrent layer and mixer. (a) Each frozen recurrent unit is followed by a mixer. The overlapping outlines represent recurrent layers stacked to form a recurrent network. (b) The mixer simply applies Linear, LayerNorm, and LeakyReLU. Blue denotes trainable parameters.}
\label{fig:frozen-recurrent-layer}
\Description{Frozen recurrent layer and mixer. (a) Each frozen recurrent unit is followed by a mixer. The overlapping outlines represent recurrent layers stacked to form a recurrent network. (b) The mixer simply applies Linear, LayerNorm, and LeakyReLU. Blue denotes trainable parameters.}
\end{figure}

Modern recurrent architectures often use nonlinear feature mixer after each recurrent unit~\citep{gu2021combining,gu2022efficiently,orvieto2023resurrecting,smith2023simplified} as well as Feed Forward Network (FFN) in transformer~\citep{vaswani2017attention,katharopoulos2020transformers,yang2024gated,de2024griffin}. In our experiments, we also adopt a similar design (Fig.~\ref{fig:frozen-recurrent-layer}) and conduct additional experiments to examine the effect of training these mixers while keeping the recurrent unit frozen.

We find that learning the mixers significantly improves performance for both Frozen LRU and Frozen GRU. As shown in Fig.~\ref{fig:mixer-training}, training the mixers increases the final evaluation return from 0.392 to 0.464 for Frozen LRU and from 0.341 to 0.386 for Frozen GRU, corresponding to relative improvements of 18.5\% and 13.2\%, respectively. Even the simple design used here (Fig.~\ref{fig:frozen-recurrent-layer}) substantially improves agent performance. These results suggest that learning the mixers contributes to the agent’s ability to adapt and use the memory provided by frozen recurrent networks. Modern large models use attention and complex nonlinear transformations to mix features~\citep{touvron2023llama,yang2025qwen3,team2025gemma}. Agents with frozen recurrent networks may also benefit from incorporating such designs into their mixers and readouts.

\section{Discussion}
Researchers often assume that recurrent RL requires training a memory model. Our experiments in Sec.\ref{sec:memorychain}, \ref{sec:popgym}, \ref{sec:mujoco} and \ref{sec:arcade} challenge this assumption. We show that agents can solve the tested POMDPs with frozen recurrent networks and nonlinear MLPs and mixers can learn to use frozen recurrent networks as memory. We believe this approach offers an alternative solution for partially observable RL and may also benefit recurrent RL applications in robotic learning.
\section{Limitations and Future Work}
Our study has several limitations. First, we do not conduct a large-scale, exhaustive ablation study, and our evaluation covers a limited set of benchmarks. Computational constraints also restrict the range of baselines considered. Future work should therefore include broader ablations, additional benchmarks, and stronger baselines to better assess the generality of our findings.
Second, our GRC analysis relies on specific assumptions and establishes its result under a relaxed ESP condition. In practice, random initialization, optimization dynamics, and finite-sample effects introduce additional uncertainty, so trained agents may not always satisfy the theoretical conditions exactly. Future work could investigate how closely these conditions hold in practical RL settings and how deviations from them affect performance.



\section{Conclusion}

We investigated whether recurrent networks must be learned in partially observable RL. 
We first introduced a simple architecture that uses frozen recurrent networks as memory modules. 
Across diverse POMDPs, these frozen memories retain information over long horizons and achieve competitive performance with substantially fewer trainable parameters. 
We further showed that visual encoders can be trained effectively through frozen recurrent memories to solve high-dimensional visual control tasks. 
Overall, our results suggest that learning the recurrent module is not always necessary and frozen recurrent networks provide a simple, strong, and parameter-efficient baseline for recurrent RL.


\subsection*{AI Use Statement}
We used \href{https://developers.openai.com/api/docs/guides/code-generation}{OpenAI Codex}
with \href{https://developers.openai.com/api/docs/models/gpt-6-astra}{GPT-6 Astra} for language polishing
and LaTeX formatting. We also used it to write plotting code
and code execution scripts. We did not retain complete prompt
and conversation records. We take full responsibility for
all the content, code, and results.
\bibliographystyle{ACM-Reference-Format}
\bibliography{references}

@misc{kubota2024reservoir,
  author = {Kubota, Tomoyuki and Imai, Yusuke and Tsunegi, Sumito and Nakajima, Kohei},
  title = {Reservoir Computing Generalized},
  year = {2024},
  doi = {10.48550/arXiv.2412.12104},
  url = {https://arxiv.org/abs/2412.12104v1},
  eprint = {2412.12104v1},
  archivePrefix = {arXiv},
  note = {Version 1},
}

@misc{bonassi2021stability,
  author = {Bonassi, Fabio and Farina, Marcello and Scattolini, Riccardo},
  title = {On the stability properties of {Gated Recurrent Units} neural networks},
  year = {2021},
  doi = {10.48550/arXiv.2011.06806},
  url = {https://arxiv.org/abs/2011.06806v6},
  eprint = {2011.06806v6},
  archivePrefix = {arXiv},
  note = {Revised manuscript, version 6},
}

@book{sutton1998reinforcement,
  author = {Sutton, Richard S. and Barto, Andrew G.},
  title = {Reinforcement Learning: An Introduction},
  year = {1998},
  publisher = {MIT Press},
  url = {https://mitpress.mit.edu/9780262193986/reinforcement-learning/},
  isbn = {9780262193986},
}

@article{verstraeten2007experimental,
  author = {Verstraeten, David and Schrauwen, Benjamin and D'Haene, Michiel and Stroobandt, Dirk},
  title = {An experimental unification of reservoir computing methods},
  year = {2007},
  journal = {Neural Networks},
  volume = {20},
  number = {3},
  pages = {391--403},
  publisher = {Elsevier},
  doi = {10.1016/j.neunet.2007.04.003},
}

@article{jaeger2004harnessing,
  author = {Jaeger, Herbert and Haas, Harald},
  title = {Harnessing nonlinearity: Predicting chaotic systems and saving energy in wireless communication},
  year = {2004},
  journal = {Science},
  volume = {304},
  number = {5667},
  pages = {78--80},
  publisher = {American Association for the Advancement of Science},
  doi = {10.1126/science.1091277},
}

@article{maass2002real,
  author = {Maass, Wolfgang and Natschl{\"a}ger, Thomas and Markram, Henry},
  title = {Real-time computing without stable states: A new framework for neural computation based on perturbations},
  year = {2002},
  journal = {Neural Computation},
  volume = {14},
  number = {11},
  pages = {2531--2560},
  publisher = {MIT Press},
}

@techreport{jaeger2001echo,
  author = {Jaeger, Herbert},
  title = {The ``echo state'' approach to analysing and training recurrent neural networks},
  year = {2001},
  number = {148},
  institution = {German National Research Center for Information Technology},
  type = {GMD Report},
  url = {https://www.ai.rug.nl/minds/uploads/EchoStatesTechRep.pdf},
  note = {Corrected version with an erratum, January 26, 2010},
}

@article{yildiz2012re,
  author = {Yildiz, Izzet B and Jaeger, Herbert and Kiebel, Stefan J},
  title = {Re-visiting the echo state property},
  year = {2012},
  journal = {Neural Networks},
  volume = {35},
  pages = {1--9},
  publisher = {Elsevier},
  doi = {10.1016/j.neunet.2012.07.005},
}

@article{gallicchio2017deep,
  author = {Gallicchio, Claudio and Micheli, Alessio and Pedrelli, Luca},
  title = {Deep reservoir computing: A critical experimental analysis},
  year = {2017},
  journal = {Neurocomputing},
  volume = {268},
  pages = {87--99},
  publisher = {Elsevier},
  doi = {10.1016/j.neucom.2016.12.089},
  url = {https://www.sciencedirect.com/science/article/pii/S0925231217307567},
}

@inproceedings{gu2022efficiently,
  author = {Gu, Albert and Goel, Karan and R{\'e}, Christopher},
  title = {Efficiently Modeling Long Sequences with Structured State Spaces},
  year = {2022},
  booktitle = {International Conference on Learning Representations},
  url = {https://openreview.net/forum?id=uYLFoz1vlAC},
}

@inproceedings{cho2014learning,
  author = {Cho, Kyunghyun and van Merri{\"e}nboer, Bart and Gulcehre, Caglar and Bahdanau, Dzmitry and Bougares, Fethi and Schwenk, Holger and Bengio, Yoshua},
  title = {Learning Phrase Representations using {RNN} Encoder--Decoder for Statistical Machine Translation},
  year = {2014},
  booktitle = {Proceedings of the 2014 Conference on Empirical Methods in Natural Language Processing (EMNLP)},
  pages = {1724--1734},
  publisher = {Association for Computational Linguistics},
  doi = {10.3115/v1/D14-1179},
  url = {https://aclanthology.org/D14-1179/},
}

@inproceedings{orvieto2023resurrecting,
  author = {Orvieto, Antonio and Smith, Samuel L and Gu, Albert and Fernando, Anushan and Gulcehre, Caglar and Pascanu, Razvan and De, Soham},
  title = {Resurrecting recurrent neural networks for long sequences},
  year = {2023},
  booktitle = {Proceedings of the 40th International Conference on Machine Learning},
  series = {Proceedings of Machine Learning Research},
  volume = {202},
  pages = {26670--26698},
  publisher = {PMLR},
  url = {https://proceedings.mlr.press/v202/orvieto23a.html},
}

@inproceedings{dong2020reservoir,
  author = {Dong, Jonathan and Ohana, Ruben and Rafayelyan, Mushegh and Krzakala, Florent},
  title = {Reservoir computing meets recurrent kernels and structured transforms},
  year = {2020},
  booktitle = {Advances in Neural Information Processing Systems},
  volume = {33},
  pages = {16785--16796},
  url = {https://proceedings.neurips.cc/paper_files/paper/2020/hash/c348616cd8a86ee661c7c98800678fad-Abstract.html},
}

@inproceedings{pinna2026paralesn,
  author = {Matteo Pinna and Giacomo Lagomarsini and Andrea Ceni and Claudio Gallicchio},
  title = {Paral{ESN}: Enabling parallel information processing in Reservoir Computing},
  year = {2026},
  booktitle = {Forty-third International Conference on Machine Learning},
  url = {https://openreview.net/forum?id=NsJUKcONFR},
}

@inproceedings{morad2023popgym,
  author = {Steven Morad and Ryan Kortvelesy and Matteo Bettini and Stephan Liwicki and Amanda Prorok},
  title = {{POPGym}: Benchmarking Partially Observable Reinforcement Learning},
  year = {2023},
  booktitle = {The Eleventh International Conference on Learning Representations},
  url = {https://openreview.net/forum?id=chDrutUTs0K},
}

@inproceedings{Osband2020Behaviour,
  author = {Ian Osband and Yotam Doron and Matteo Hessel and John Aslanides and Eren Sezener and Andre Saraiva and Katrina McKinney and Tor Lattimore and Csaba Szepesvari and Satinder Singh and Benjamin Van Roy and Richard Sutton and David Silver and Hado Van Hasselt},
  title = {Behaviour Suite for Reinforcement Learning},
  year = {2020},
  booktitle = {International Conference on Learning Representations},
  url = {https://openreview.net/forum?id=rygf-kSYwH},
}

@software{gymnax2022github,
  author = {Robert Tjarko Lange},
  title = {{gymnax}: A {JAX}-based Reinforcement Learning Environment Library},
  year = {2022},
  version = {0.0.4},
  url = {https://github.com/RobertTLange/gymnax},
}

@inproceedings{smith2023simplified,
  author = {Jimmy T.H. Smith and Andrew Warrington and Scott Linderman},
  title = {Simplified State Space Layers for Sequence Modeling},
  year = {2023},
  booktitle = {The Eleventh International Conference on Learning Representations},
  url = {https://openreview.net/forum?id=Ai8Hw3AXqks},
}

@inproceedings{wang2026investigating,
  author = {Zekang Wang and Zhe He and Borong Zhang and Edan Toledo and Steven Morad},
  title = {Investigating Memory in Model-Free {RL} with {POPG}ym Arcade},
  year = {2026},
  booktitle = {Proceedings of the 43rd International Conference on Machine Learning},
  series = {Proceedings of Machine Learning Research},
  volume = {306},
  pages = {126692--126733},
  publisher = {PMLR},
  url = {https://proceedings.mlr.press/v306/wang26bx.html},
}

@inproceedings{lu2023structured,
  author = {Chris Lu and Yannick Schroecker and Albert Gu and Emilio Parisotto and Jakob Nicolaus Foerster and Satinder Singh and Feryal Behbahani},
  title = {Structured State Space Models for In-Context Reinforcement Learning},
  year = {2023},
  booktitle = {Advances in Neural Information Processing Systems},
  volume = {36},
  pages = {47016--47031},
  doi = {10.52202/075280-2036},
  url = {https://papers.neurips.cc/paper_files/paper/2023/hash/92d3d2a9801211ca3693ccb2faa1316f-Abstract-Conference.html},
}

@article{elman1990finding,
  author = {Elman, Jeffrey L.},
  title = {Finding structure in time},
  year = {1990},
  journal = {Cognitive Science},
  volume = {14},
  number = {2},
  pages = {179--211},
  publisher = {Wiley},
  doi = {10.1207/s15516709cog1402_1},
}

@article{kaelbling1998planning,
  author = {Kaelbling, Leslie Pack and Littman, Michael L and Cassandra, Anthony R},
  title = {Planning and acting in partially observable stochastic domains},
  year = {1998},
  journal = {Artificial Intelligence},
  volume = {101},
  number = {1--2},
  pages = {99--134},
  publisher = {Elsevier},
  doi = {10.1016/S0004-3702(98)00023-X},
}

@inproceedings{hausknecht2015deep,
  author = {Hausknecht, Matthew and Stone, Peter},
  title = {Deep Recurrent Q-Learning for Partially Observable {MDP}s},
  year = {2015},
  booktitle = {AAAI Fall Symposium on Sequential Decision Making for Intelligent Agents},
  url = {https://aaai.org/papers/11673-11673-deep-recurrent-q-learning-for-partially-observable-mdps/},
}

@inproceedings{ni2022recurrent,
  author = {Ni, Tianwei and Eysenbach, Benjamin and Salakhutdinov, Ruslan},
  title = {Recurrent Model-Free {RL} Can Be a Strong Baseline for Many {POMDP}s},
  year = {2022},
  booktitle = {Proceedings of the 39th International Conference on Machine Learning},
  series = {Proceedings of Machine Learning Research},
  volume = {162},
  pages = {16691--16723},
  publisher = {PMLR},
  url = {https://proceedings.mlr.press/v162/ni22a.html},
}

@inproceedings{szita2006reinforcement,
  author = {Szita, Istv{\'a}n and Gyenes, Viktor and L{\H{o}}rincz, Andr{\'a}s},
  title = {Reinforcement Learning with Echo State Networks},
  year = {2006},
  booktitle = {Artificial Neural Networks---ICANN 2006},
  series = {Lecture Notes in Computer Science},
  volume = {4131},
  pages = {830--839},
  publisher = {Springer},
  doi = {10.1007/11840817_86},
}

@article{yamashita2022reinforcement,
  author = {Yamashita, Kodai and Hamagami, Tomoki},
  title = {Reinforcement Learning for {POMDP} Environments Using State Representation with Reservoir Computing},
  year = {2022},
  journal = {Journal of Advanced Computational Intelligence and Intelligent Informatics},
  volume = {26},
  number = {4},
  pages = {562--569},
  doi = {10.20965/jaciii.2022.p0562},
}

@misc{mckee2024reservoir,
  author = {McKee, Kevin},
  title = {Reservoir Computing for Fast, Simplified Reinforcement Learning on Memory Tasks},
  year = {2024},
  doi = {10.48550/arXiv.2412.13093},
  url = {https://arxiv.org/abs/2412.13093},
  eprint = {2412.13093},
  archivePrefix = {arXiv},
}

@article{hochreiter1997long,
  author = {Hochreiter, Sepp and Schmidhuber, J{\"u}rgen},
  title = {Long Short-Term Memory},
  year = {1997},
  journal = {Neural Computation},
  volume = {9},
  number = {8},
  pages = {1735--1780},
  publisher = {MIT Press},
  doi = {10.1162/neco.1997.9.8.1735},
}

@inproceedings{vaswani2017attention,
  author = {Vaswani, Ashish and Shazeer, Noam and Parmar, Niki and Uszkoreit, Jakob and Jones, Llion and Gomez, Aidan N and Kaiser, {\L}ukasz and Polosukhin, Illia},
  title = {Attention Is All You Need},
  year = {2017},
  booktitle = {Advances in Neural Information Processing Systems},
  volume = {30},
  url = {https://papers.nips.cc/paper_files/paper/2017/hash/3f5ee243547dee91fbd053c1c4a845aa-Abstract.html},
}

@inproceedings{katharopoulos2020transformers,
  author = {Katharopoulos, Angelos and Vyas, Apoorv and Pappas, Nikolaos and Fleuret, Fran{\c{c}}ois},
  title = {Transformers are {RNN}s: Fast Autoregressive Transformers with Linear Attention},
  year = {2020},
  booktitle = {Proceedings of the 37th International Conference on Machine Learning},
  series = {Proceedings of Machine Learning Research},
  volume = {119},
  pages = {5156--5165},
  publisher = {PMLR},
  url = {https://proceedings.mlr.press/v119/katharopoulos20a.html},
}

@inproceedings{gu2024mamba,
  author = {Albert Gu and Tri Dao},
  title = {{Mamba}: Linear-Time Sequence Modeling with Selective State Spaces},
  year = {2024},
  booktitle = {First Conference on Language Modeling},
  url = {https://openreview.net/forum?id=tEYskw1VY2},
}

@misc{schulman2017proximal,
  author = {Schulman, John and Wolski, Filip and Dhariwal, Prafulla and Radford, Alec and Klimov, Oleg},
  title = {Proximal policy optimization algorithms},
  year = {2017},
  url = {https://arxiv.org/abs/1707.06347},
  eprint = {1707.06347},
  archivePrefix = {arXiv},
}

@inproceedings{gallici2025simplifying,
  author = {Gallici, Matteo and Fellows, Mattie and Ellis, Benjamin and Pou, Bartomeu and Masmitja, Ivan and Foerster, Jakob Nicolaus and Martin, Mario},
  title = {Simplifying Deep Temporal Difference Learning},
  year = {2025},
  booktitle = {The Thirteenth International Conference on Learning Representations},
  url = {https://proceedings.iclr.cc/paper_files/paper/2025/hash/c23f3852601f6dd7f0b39223d031806f-Abstract-Conference.html},
}

@inproceedings{kapturowski2019recurrent,
  author = {Kapturowski, Steven and Ostrovski, Georg and Quan, John and Munos, R{\'e}mi and Dabney, Will},
  title = {Recurrent Experience Replay in Distributed Reinforcement Learning},
  year = {2019},
  booktitle = {International Conference on Learning Representations},
  url = {https://openreview.net/forum?id=r1lyTjAqYX},
}

@inproceedings{parisotto2020stabilizing,
  author = {Parisotto, Emilio and Song, Francis and Rae, Jack and Pascanu, Razvan and Gulcehre, Caglar and Jayakumar, Siddhant and Jaderberg, Max and Kaufman, Rapha{\"e}l Lopez and Clark, Aidan and Noury, Seb and Botvinick, Matthew and Heess, Nicolas and Hadsell, Raia},
  title = {Stabilizing Transformers for Reinforcement Learning},
  year = {2020},
  booktitle = {Proceedings of the 37th International Conference on Machine Learning},
  series = {Proceedings of Machine Learning Research},
  volume = {119},
  pages = {7487--7498},
  publisher = {PMLR},
  url = {https://proceedings.mlr.press/v119/parisotto20a.html},
}

@inproceedings{gu2022parameterization,
  author = {Gu, Albert and Gupta, Ankit and Goel, Karan and R{\'e}, Christopher},
  title = {On the Parameterization and Initialization of Diagonal State Space Models},
  year = {2022},
  booktitle = {Advances in Neural Information Processing Systems},
  volume = {35},
  pages = {35971--35983},
  doi = {10.52202/068431-2607},
  url = {https://proceedings.neurips.cc/paper_files/paper/2022/hash/e9a32fade47b906de908431991440f7c-Abstract-Conference.html},
}

@inproceedings{lu2024rethinking,
  author = {Lu, Chenhao and Shi, Ruizhe and Liu, Yuyao and Hu, Kaizhe and Du, Simon Shaolei and Xu, Huazhe},
  title = {Rethinking Transformers in Solving {POMDP}s},
  year = {2024},
  booktitle = {Proceedings of the 41st International Conference on Machine Learning},
  series = {Proceedings of Machine Learning Research},
  volume = {235},
  pages = {33089--33112},
  publisher = {PMLR},
  url = {https://proceedings.mlr.press/v235/lu24h.html},
}

@inproceedings{ni2023when,
  author = {Ni, Tianwei and Ma, Michel and Eysenbach, Benjamin and Bacon, Pierre-Luc},
  title = {When Do Transformers Shine in {RL}? Decoupling Memory from Credit Assignment},
  year = {2023},
  booktitle = {Advances in Neural Information Processing Systems},
  volume = {36},
  pages = {50429--50452},
  doi = {10.52202/075280-2194},
  url = {https://papers.nips.cc/paper_files/paper/2023/hash/9dc5accb1e4f4a9798eae145f2e4869b-Abstract-Conference.html},
}

@inproceedings{morad2024memoroids,
  author = {Morad, Steven and Lu, Chris and Kortvelesy, Ryan and Liwicki, Stephan and Foerster, Jakob and Prorok, Amanda},
  title = {Recurrent Reinforcement Learning with Memoroids},
  year = {2024},
  booktitle = {Advances in Neural Information Processing Systems},
  volume = {37},
  pages = {14386--14416},
  doi = {10.52202/079017-0459},
  url = {https://proceedings.neurips.cc/paper_files/paper/2024/hash/19f7f755908372efb25826d61959cdf9-Abstract-Conference.html},
}

@inproceedings{zakka2025playground,
  author = {Zakka, Kevin and Tabanpour, Baruch and Liao, Qiayuan and Haiderbhai, Mustafa and Holt, Samuel and Luo, Jing Yuan and Allshire, Arthur and Frey, Erik and Sreenath, Koushil and Kahrs, Lueder Alexander and Sferrazza, Carmelo and Tassa, Yuval and Abbeel, Pieter},
  title = {Demonstrating {MuJoCo Playground}},
  year = {2025},
  booktitle = {Proceedings of Robotics: Science and Systems},
  doi = {10.15607/RSS.2025.XXI.020},
  url = {https://www.roboticsproceedings.org/rss21/p020.html},
}

@inproceedings{banino2022coberl,
  author = {Andrea Banino and Adria Puigdomenech Badia and Jacob C Walker and Tim Scholtes and Jovana Mitrovic and Charles Blundell},
  title = {Co{BERL}: Contrastive {BERT} for Reinforcement Learning},
  year = {2022},
  booktitle = {International Conference on Learning Representations},
  url = {https://openreview.net/forum?id=sRZ3GhmegS},
}

@article{pramanik2024agalite,
  author = {Subhojeet Pramanik and Esraa Elelimy and Marlos C. Machado and Adam White},
  title = {{AGaLiTe}: Approximate Gated Linear Transformers for Online Reinforcement Learning},
  year = {2024},
  journal = {Transactions on Machine Learning Research},
  url = {https://openreview.net/forum?id=lh6vOAHuvo},
  issn = {2835-8856},
}

@article{gallicchio2018design,
  author = {Gallicchio, Claudio and Micheli, Alessio and Pedrelli, Luca},
  title = {Design of deep echo state networks},
  year = {2018},
  journal = {Neural Networks},
  volume = {108},
  pages = {33--47},
  publisher = {Elsevier},
  doi = {10.1016/j.neunet.2018.08.002},
  url = {https://www.sciencedirect.com/science/article/pii/S0893608018302223},
}

@misc{ba2016layer,
  author = {Ba, Jimmy Lei and Kiros, Jamie Ryan and Hinton, Geoffrey E},
  title = {Layer normalization},
  year = {2016},
  url = {https://arxiv.org/abs/1607.06450},
  eprint = {1607.06450},
  archivePrefix = {arXiv},
}

@software{jax2018github,
  author = {James Bradbury and Roy Frostig and Peter Hawkins and Matthew James Johnson and Yash Katariya and Chris Leary and Dougal Maclaurin and George Necula and Adam Paszke and Jake Vander{P}las and Skye Wanderman-{M}ilne and Qiao Zhang},
  title = {{JAX}: composable transformations of {P}ython+{N}um{P}y programs},
  year = {2018},
  version = {0.3.13},
  url = {https://github.com/jax-ml/jax},
}

@article{kidger2021equinox,
  author = {Patrick Kidger and Cristian Garcia},
  title = {{E}quinox: neural networks in {JAX} via callable {P}y{T}rees and filtered transformations},
  year = {2021},
  journal = {Differentiable Programming workshop at Neural Information Processing Systems 2021},
  url = {https://arxiv.org/abs/2111.00254},
}

@inproceedings{meng2021memory,
  author = {Meng, Lingheng and Gorbet, Rob and Kuli{\'c}, Dana},
  title = {Memory-based Deep Reinforcement Learning for {POMDPs}},
  year = {2021},
  booktitle = {2021 IEEE/RSJ International Conference on Intelligent Robots and Systems (IROS)},
  pages = {5619--5626},
  organization = {IEEE},
  doi = {10.1109/IROS51168.2021.9636140},
  url = {https://arxiv.org/abs/2102.12344},
}

@misc{tallec2018reproducing,
  author = {Tallec, Corentin and Blier, L{\'e}onard and Kalainathan, Diviyan},
  title = {Reproducing ``{World Models}'': Is training the recurrent network really needed?},
  year = {2018},
  url = {https://ctallec.github.io/world-models/},
  note = {Accessed: 2026-10-08},
  howpublished = {Blog post},
}

@inproceedings{NEURIPS2018_2de5d166,
  author = {Ha, David and Schmidhuber, J\"{u}rgen},
  editor = {S. Bengio and H. Wallach and H. Larochelle and K. Grauman and N. Cesa-Bianchi and R. Garnett},
  title = {Recurrent World Models Facilitate Policy Evolution},
  year = {2018},
  booktitle = {Advances in Neural Information Processing Systems},
  volume = {31},
  publisher = {Curran Associates, Inc.},
  url = {https://proceedings.neurips.cc/paper/2018/hash/2de5d16682c3c35007e4e92982f1a2ba-Abstract.html},
}

@article{lukovsevivcius2009reservoir,
  author = {Luko{\v{s}}evi{\v{c}}ius, Mantas and Jaeger, Herbert},
  title = {Reservoir computing approaches to recurrent neural network training},
  year = {2009},
  journal = {Computer Science Review},
  volume = {3},
  number = {3},
  pages = {127--149},
  publisher = {Elsevier},
  doi = {10.1016/j.cosrev.2009.03.005},
  url = {https://www.sciencedirect.com/science/article/pii/S1574013709000173},
}

@inproceedings{todorov2012mujoco,
  author = {Todorov, Emanuel and Erez, Tom and Tassa, Yuval},
  title = {{MuJoCo}: A Physics Engine for Model-based Control},
  year = {2012},
  booktitle = {2012 IEEE/RSJ International Conference on Intelligent Robots and Systems},
  pages = {5026--5033},
  organization = {IEEE},
  doi = {10.1109/IROS.2012.6386109},
  url = {https://ieeexplore.ieee.org/document/6386109/},
}

@inproceedings{gu2021combining,
  author = {Gu, Albert and Johnson, Isys and Goel, Karan and Saab, Khaled and Dao, Tri and Rudra, Atri and R{\'e}, Christopher},
  title = {Combining recurrent, convolutional, and continuous-time models with linear state space layers},
  year = {2021},
  booktitle = {Advances in Neural Information Processing Systems},
  volume = {34},
  pages = {572--585},
  url = {https://proceedings.neurips.cc/paper/2021/hash/05546b0e38ab9175cd905eebcc6ebb76-Abstract.html},
}

@inproceedings{yang2024gated,
  author = {Songlin Yang and Bailin Wang and Yikang Shen and Rameswar Panda and Yoon Kim},
  title = {Gated Linear Attention Transformers with Hardware-Efficient Training},
  year = {2024},
  booktitle = {Proceedings of the 41st International Conference on Machine Learning},
  series = {Proceedings of Machine Learning Research},
  volume = {235},
  pages = {56501--56523},
  publisher = {PMLR},
  url = {https://proceedings.mlr.press/v235/yang24ab.html},
}

@misc{de2024griffin,
  author = {De, Soham and Smith, Samuel L and Fernando, Anushan and Botev, Aleksandar and Cristian-Muraru, George and Gu, Albert and Haroun, Ruba and Berrada, Leonard and Chen, Yutian and Srinivasan, Srivatsan and others},
  title = {{Griffin}: Mixing Gated Linear Recurrences with Local Attention for Efficient Language Models},
  year = {2024},
  url = {https://arxiv.org/abs/2402.19427},
  eprint = {2402.19427},
  archivePrefix = {arXiv},
}

@misc{touvron2023llama,
  author = {Touvron, Hugo and Lavril, Thibaut and Izacard, Gautier and Martinet, Xavier and Lachaux, Marie-Anne and Lacroix, Timoth{\'e}e and Rozi{\`e}re, Baptiste and Goyal, Naman and Hambro, Eric and Azhar, Faisal and others},
  title = {{LLaMA}: Open and Efficient Foundation Language Models},
  year = {2023},
  url = {https://arxiv.org/abs/2302.13971},
  eprint = {2302.13971},
  archivePrefix = {arXiv},
}

@misc{yang2025qwen3,
  author = {Yang, An and Li, Anfeng and Yang, Baosong and Zhang, Beichen and Hui, Binyuan and Zheng, Bo and Yu, Bowen and Gao, Chang and Huang, Chengen and Lv, Chenxu and others},
  title = {{Qwen3} Technical Report},
  year = {2025},
  url = {https://arxiv.org/abs/2505.09388},
  eprint = {2505.09388},
  archivePrefix = {arXiv},
}

@misc{team2025gemma,
  author = {{Gemma Team} and Kamath, Aishwarya and Ferret, Johan and Pathak, Shreya and Vieillard, Nino and Merhej, Ramona and Perrin, Sarah and Matejovicova, Tatiana and Ram{\'e}, Alexandre and Rivi{\`e}re, Morgane and others},
  title = {{Gemma 3} Technical Report},
  year = {2025},
  url = {https://arxiv.org/abs/2503.19786},
  eprint = {2503.19786},
  archivePrefix = {arXiv},
}

@inproceedings{eberhard2025memory,
  author = {Eberhard, Onno and Muehlebach, Michael and Vernade, Claire},
  title = {Partially Observable Reinforcement Learning with Memory Traces},
  year = {2025},
  booktitle = {Proceedings of the 42nd International Conference on Machine Learning},
  series = {Proceedings of Machine Learning Research},
  volume = {267},
  pages = {14934--14949},
  publisher = {PMLR},
  url = {https://proceedings.mlr.press/v267/eberhard25a.html},
}

@inproceedings{deverett2019interval,
  author = {Deverett, Ben and Faulkner, Ryan and Fortunato, Meire and Wayne, Gregory and Leibo, Joel Z.},
  title = {Interval Timing in Deep Reinforcement Learning Agents},
  year = {2019},
  booktitle = {Advances in Neural Information Processing Systems},
  volume = {32},
  url = {https://papers.neurips.cc/paper_files/paper/2019/hash/2bf283c05b601f21364d052ca0ec798d-Abstract.html},
}

@inproceedings{bohm2023feature,
  author = {B{\"o}hm, Peter and Pounds, Pauline and Chapman, Archie C.},
  title = {Feature Extraction for Effective and Efficient Deep Reinforcement Learning on Real Robotic Platforms},
  year = {2023},
  booktitle = {2023 IEEE International Conference on Robotics and Automation (ICRA)},
  pages = {7126--7132},
  publisher = {IEEE},
  doi = {10.1109/ICRA48891.2023.10160862},
  url = {https://ieeexplore.ieee.org/document/10160862/},
}

@inproceedings{mochizuki2023characterizing,
  author = {Mochizuki-Freeman, James and Maini, Sahaj Singh and Tiganj, Zoran},
  title = {Characterizing Neural Activity in Cognitively Inspired {RL} Agents during an Evidence Accumulation Task},
  year = {2023},
  booktitle = {2023 International Joint Conference on Neural Networks (IJCNN)},
  publisher = {IEEE},
  doi = {10.1109/IJCNN54540.2023.10190998},
  url = {https://homes.luddy.indiana.edu/ztiganj/papers/EvidenceAccumulationRL_IJCNN2023.pdf},
}

@inproceedings{elelimy2024realtime,
  author = {Elelimy, Esraa and White, Adam and Bowling, Michael and White, Martha},
  title = {Real-Time Recurrent Learning using Trace Units in Reinforcement Learning},
  booktitle = {Advances in Neural Information Processing Systems},
  volume = {37},
  year = {2024},
  doi = {10.52202/079017-0541},
  url = {https://papers.nips.cc/paper_files/paper/2024/hash/1e616bde0438cb10cb6adf076ae7d336-Abstract-Conference.html},
}

@inproceedings{lampinen2021towards,
  author = {Lampinen, Andrew K. and Chan, Stephanie C. Y. and Banino, Andrea and Hill, Felix},
  title = {Towards mental time travel: a hierarchical memory for reinforcement learning agents},
  booktitle = {Advances in Neural Information Processing Systems},
  volume = {34},
  year = {2021},
  url = {https://papers.nips.cc/paper/2021/hash/ed519dacc89b2bead3f453b0b05a4a8b-Abstract.html},
}

@misc{esslinger2022deep,
  author = {Esslinger, Kevin and Platt, Robert and Amato, Christopher},
  title = {Deep Transformer {Q}-Networks for Partially Observable Reinforcement Learning},
  year = {2022},
  eprint = {2206.01078},
  archivePrefix = {arXiv},
  primaryClass = {cs.LG},
  doi = {10.48550/arXiv.2206.01078},
  url = {https://arxiv.org/abs/2206.01078},
}

@inproceedings{morad2023reinforcement,
  author = {Morad, Steven and Kortvelesy, Ryan and Liwicki, Stephan and Prorok, Amanda},
  title = {Reinforcement Learning with Fast and Forgetful Memory},
  booktitle = {Advances in Neural Information Processing Systems},
  volume = {36},
  year = {2023},
  doi = {10.52202/075280-3150},
  url = {https://proceedings.neurips.cc/paper_files/paper/2023/hash/e3bf2f0f10774c474de22a12cb060e2c-Abstract-Conference.html},
}

@inproceedings{le2025stable,
  author = {Le, Hung and Nguyen, Dung and Do, Kien and Gupta, Sunil and Venkatesh, Svetha},
  title = {Stable Hadamard Memory: Revitalizing Memory-Augmented Agents for Reinforcement Learning},
  booktitle = {The Thirteenth International Conference on Learning Representations},
  year = {2025},
  url = {https://openreview.net/forum?id=We5z3UEnUY},
}

@inproceedings{cherepanov2026recurrent,
  author = {Cherepanov, Egor and Staroverov, Aleksei and Kovalev, Alexey K. and Panov, Aleksandr I.},
  title = {Recurrent Action Transformer with Memory},
  booktitle = {The Fourteenth International Conference on Learning Representations},
  year = {2026},
  url = {https://openreview.net/forum?id=kByN4v0M3e},
}

@misc{wayne2018unsupervised,
  author = {Wayne, Greg and Hung, Chia-Chun and Amos, David and Mirza, Mehdi and Ahuja, Arun and Grabska-Barwinska, Agnieszka and Rae, Jack and Mirowski, Piotr and Leibo, Joel Z. and Santoro, Adam and Gemici, Mevlana and Reynolds, Malcolm and Harley, Tim and Abramson, Josh and Mohamed, Shakir and Rezende, Danilo and Saxton, David and Cain, Adam and Hillier, Chloe and Silver, David and Kavukcuoglu, Koray and Botvinick, Matt and Hassabis, Demis and Lillicrap, Timothy},
  title = {Unsupervised Predictive Memory in a Goal-Directed Agent},
  year = {2018},
  eprint = {1803.10760},
  archivePrefix = {arXiv},
  primaryClass = {cs.LG},
  doi = {10.48550/arXiv.1803.10760},
  url = {https://arxiv.org/abs/1803.10760},
}

@inproceedings{fortunato2019generalization,
  author = {Fortunato, Meire and Tan, Melissa and Faulkner, Ryan and Hansen, Steven and Puigdom{\`e}nech Badia, Adri{\`a} and Buttimore, Gavin and Deck, Charles and Leibo, Joel Z. and Blundell, Charles},
  title = {Generalization of Reinforcement Learners with Working and Episodic Memory},
  booktitle = {Advances in Neural Information Processing Systems},
  volume = {32},
  year = {2019},
  url = {https://proceedings.neurips.cc/paper/2019/hash/02ed812220b0705fabb868ddbf17ea20-Abstract.html},
}

\clearpage
\appendix
\setcounter{figure}{0}
\setcounter{table}{0}
\setcounter{equation}{0}
\renewcommand{\thefigure}{A\arabic{figure}}
\renewcommand{\thetable}{A\arabic{table}}
\renewcommand{\theequation}{A\arabic{equation}}
\renewcommand{\theHfigure}{appendix.\arabic{figure}}
\renewcommand{\theHtable}{appendix.\arabic{table}}
\renewcommand{\theHequation}{appendix.\arabic{equation}}

\twocolumn[{%
\begin{minipage}{\textwidth}
\begin{center}
\Huge\sffamily\bfseries
Supplementary Material
\end{center}
\centering
\resizebox{\textwidth}{!}{
\begin{tikzpicture}[x=1cm,y=1.25cm,
  every node/.style={font=\sffamily\normalsize},
  block/.style={draw=black!45,line width=.65pt,rounded corners=2pt,
    align=center,minimum height=1.10cm,text width=2.6cm,inner sep=5pt,fill=white},
  fixed/.style={block,fill=black!9},
  learned/.style={block,draw=blue!55!black,fill=blue!7},
  flow/.style={-{Triangle[length=2mm,width=1.3mm]},line width=.85pt,
    draw=black!70,shorten >=1pt},
  update/.style={-{Triangle[length=2mm,width=1.3mm]},dashed,
    line width=1pt,draw=blue!65!black,shorten >=1pt},
  note/.style={font=\sffamily\normalsize,
    text=black!85,inner sep=2pt},
  update note/.style={note,text=blue!65!black},
  legend text/.style={anchor=west,font=\sffamily\normalsize,inner sep=0pt},
  layer/.style={block,minimum height=.52cm,text width=2.15cm,inner sep=3pt},
  learned layer/.style={layer,draw=blue!55!black,fill=blue!7},
  layer flow/.style={-{Triangle[length=1.4mm,width=1mm]},line width=.8pt,
    draw=black!70,shorten <=.5pt,shorten >=.5pt}]

\node[block,text width=1.95cm] (obs) at (0,0)
  {\textbf{Observation}\\$o_t$};
\node[learned,text width=2.4cm] (cnn) at (3.55,0)
  {\textbf{Trainable CNN}\\$E_{\psi}$\quad{\footnotesize(Optional)}};
\node[fixed,text width=3.6cm] (mem) at (7.85,0)
  {\textbf{Frozen Recurrent\\Network}\\$h_t=f_{\omega_0}(h_{t-1},u_t)$};
\node[learned] (read) at (12.25,0)
  {\textbf{Actor / Critic MLPs}\\$\pi_{\theta}(a_t\mid h_t)$\\$V_{\phi}(h_t)$};
\node[block,text width=2.1cm] (env) at (16.1,0)
  {\textbf{Environment}};

\draw[flow] (obs)--node[note,below] {Images}(cnn);
\draw[flow] (cnn)--node[note,below] {$u_t$}(mem);
\draw[flow] (mem)--node[note,above] {$h_t$}(read);
\draw[flow] (read)--node[note,above] {$a_t$}(env);

\draw[flow,rounded corners=2pt] (1.55,0) -- (1.55,.98)
  -- node[note,above] {Vectors} (5.52,.98) -- (5.52,0);
\draw[flow,rounded corners=2pt] (env.north) -- (16.1,1.68)
  -- node[note,above] {Observation Feedback} (0,1.68) -- (obs.north);

\node[block] (loss) at (12.25,-2.4)
  {\textbf{PPO Learning}\\Policy + Value Loss\\$-$ Entropy Bonus};
\draw[flow] ([xshift=-5.5mm]read.south)
  -- node[note,left] {Predictions} ([xshift=-5.5mm]loss.north);
\draw[flow,rounded corners=2pt] (env.south) -- (16.1,-2.4)
  -- node[note,above] {Rollout / GAE} (loss.east);
\draw[update] ([xshift=5.5mm]loss.north)
  -- node[update note,right] {Update $\theta,\phi$} ([xshift=5.5mm]read.south);
\draw[update,rounded corners=2pt] (loss.west) -- (9.75,-2.4)
  -- (9.75,-1.18) -- node[update note,above] {Update $\psi$ (Images)}
  (3.55,-1.18) -- (cnn.south);

\begin{scope}[shift={(0,-1.57)}]
\draw[draw=blue!55!black,fill=blue!7,line width=.65pt,rounded corners=1pt]
  (0,-.12) rectangle (.43,.12);
\node[legend text] at (.60,0) {Trainable Parameters};
\draw[draw=black!45,fill=black!9,line width=.65pt,rounded corners=1pt]
  (0,-.60) rectangle (.43,-.36);
\node[legend text] at (.60,-.48) {Fixed Parameters};
\draw[draw=black!45,fill=white,line width=.65pt,rounded corners=1pt]
  (0,-1.08) rectangle (.43,-.84);
\node[legend text] at (.60,-.96) {Parameter-Free Elements};
\draw[flow] (5.30,-.24)--(5.96,-.24);
\node[legend text] at (6.20,-.24) {Data Flow};
\draw[update] (5.30,-.72)--(5.96,-.72);
\node[legend text] at (6.20,-.72) {Parameter Update};
\end{scope}

\draw[black!25,line width=.55pt] (18.05,2.30) -- (18.05,-3.40);
\node[font=\sffamily\normalsize\bfseries] at (19.9,2.10) {Each MLP Head};
\node[inner sep=1pt] (readinput) at (19.9,1.47) {$h_t$};
\node[learned layer] (hiddenlinear) at (19.9,.66) {\textbf{Linear Layer}};
\node[learned layer] (hiddennorm) at (19.9,-.26) {\textbf{LayerNorm}};
\node[layer] (hiddenact) at (19.9,-1.18) {\textbf{LeakyReLU}};
\draw[layer flow] (readinput.south) -- (hiddenlinear.north);
\draw[layer flow] (hiddenlinear.south) -- (hiddennorm.north);
\draw[layer flow] (hiddennorm.south) -- (hiddenact.north);
\draw[black!55,line width=.65pt]
  (21.22,.95) -- (21.42,.95) -- (21.42,-1.47) -- (21.22,-1.47);
\node[anchor=west,inner sep=2pt] at (21.44,-.26) {$\times K$};
\node[learned layer,minimum height=.68cm] (outputlinear) at (19.9,-2.23)
  {\textbf{Linear Layer}\\Output};
\node[inner sep=1pt,align=center,text width=3.2cm] (readoutput) at (19.9,-3.15)
  {Actor: Logits Or Mean\\Critic: Scalar Value};
\draw[layer flow] (hiddenact.south) -- (outputlinear.north);
\draw[layer flow] (outputlinear.south) -- (readoutput.north);
\end{tikzpicture}
}

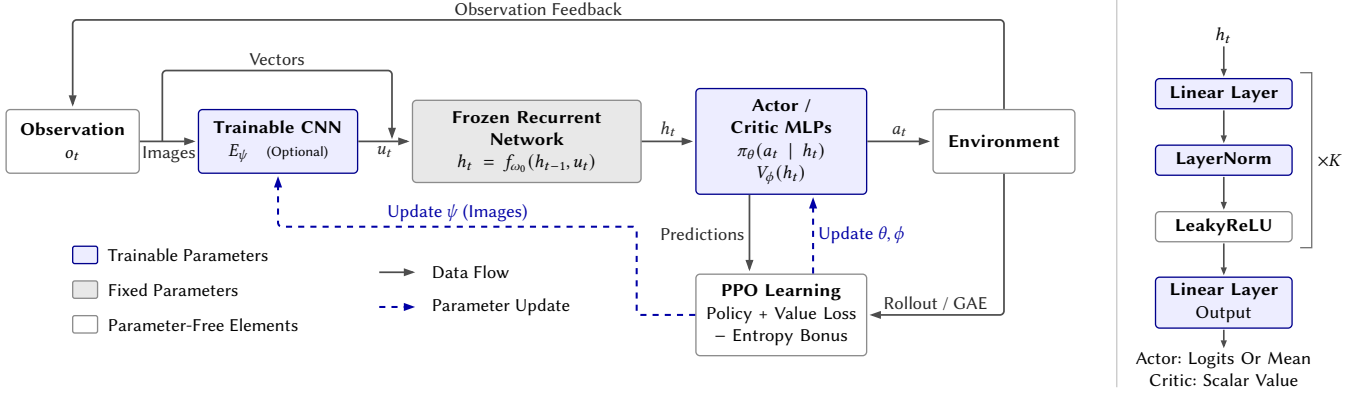
\captionof{figure}{Overview of recurrent PPO agent with frozen recurrent networks. The frozen recurrent network supplies memory features to trainable actor and critic MLPs, which are optimized with PPO.}
\label{fig:app-ppo}
\Description{Observations reach the frozen recurrent network directly for vectors or through an optional trainable CNN for images. Memory features feed separate actor and critic MLPs. The environment supplies observations and rollout data, and PPO updates the trainable parameters. The right panel shows the layers in each MLP head.}
\end{minipage}
\vspace{1em}
}]

\section{PPO Agent}
\label{app:ppo}
We show the recurrent PPO agent used in our MuJoCo Playground experiments in Fig.~\ref{fig:app-ppo}.
It follows the PQN architecture in Fig.~\ref{fig:hero}, with separate actor and critic MLPs in place of the PQN readout.
The actor produces the policy, and the critic estimates the state value.
We update the trainable parameters using PPO~\citep{schulman2017proximal}.

\section{Deep Reservoir Computing}
\label{app:variants}
Deep Reservoir Computing (Deep RC)~\citep{gallicchio2017deep} stacks reservoir layers to construct temporal features from input history and uses a trainable readout to learn from their hidden states.
To further validate our conclusions, we conduct additional experiments with three Deep RC models: Deep Echo State Network (DeepESN), Structured Reservoir Computing (SRC), and Parallel Echo State Network (ParalESN).

\subsection{Deep Echo State Network (DeepESN)}
\label{app:deepesn}
DeepESN extends a standard ESN by stacking multiple reservoir layers~\citep{gallicchio2018design}.
Let $\mathbf h_t^{(\ell)}$ denote the state of layer $\ell$ at time $t$.
The first layer receives $\mathbf u_t^{(1)}=o_t$, while each higher layer receives $\mathbf u_t^{(\ell)}=\mathbf h_t^{(\ell-1)}$.
Each layer computes a candidate state and applies a leaky update:
\begin{equation}
\begin{aligned}
\widetilde{\mathbf h}_t^{(\ell)}
&=\tanh\!\left(\mathbf W_{\mathrm{in}}^{(\ell)}\mathbf u_t^{(\ell)}
+\mathbf W_{\mathrm{rec}}^{(\ell)}\mathbf h_{t-1}^{(\ell)}+\mathbf b^{(\ell)}\right),\\
\mathbf h_t^{(\ell)}
&=(1-\alpha^{(\ell)})\mathbf h_{t-1}^{(\ell)}
+\alpha^{(\ell)}\widetilde{\mathbf h}_t^{(\ell)}.
\end{aligned}
\label{eq:app-deepesn}
\end{equation}
Here, $\mathbf W_{\mathrm{in}}^{(\ell)}$ and $\mathbf W_{\mathrm{rec}}^{(\ell)}$ are the fixed input and recurrent weight matrices, and $\mathbf b^{(\ell)}$ is a fixed bias.
The leak rate $\alpha^{(\ell)}\in(0,1]$ controls how much of the candidate replaces the previous state.
For $L_{\mathrm r}$ reservoir layers, a typical readout uses
\begin{equation}
\mathbf y_t=R_\theta\!\left([\mathbf h_t^{(1)};\ldots;\mathbf h_t^{(L_{\mathrm r})}]\right),
\label{eq:app-deepesn-readout}
\end{equation}
where $[\,\cdot\,;\,\cdot\,]$ denotes concatenation and $R_\theta$ is the trainable readout.
This gives the readout access to temporal features at different depths.

\subsection{Structured Reservoir Computing (SRC)}
\label{app:src}
SRC replaces dense random projections with structured random transforms~\citep{dong2020reservoir}.
For a layer with $N$ state coordinates and input $\mathbf u_t\in\mathbb R^d$, the basic update is
\begin{equation}
\mathbf h_t=\frac{1}{\sqrt N}\phi\!\left(
\mathbf S[\sigma_{\mathrm r}\mathbf h_{t-1};\sigma_{\mathrm{in}}\mathbf u_t]\right).
\label{eq:app-src-update}
\end{equation}
Here, $\phi$ is an elementwise nonlinear activation, and $\sigma_{\mathrm r}$ and $\sigma_{\mathrm{in}}$ scale the previous state and current input.
The structured projection $\mathbf S\in\mathbb R^{N\times(N+d)}$ is constructed from a square transform of the form
\begin{equation}
\mathbf T_m=\gamma\sqrt m\,
\mathbf H_m\mathbf D_1\mathbf H_m\mathbf D_2\mathbf H_m\mathbf D_3,
\label{eq:app-src-projection}
\end{equation}
where $\mathbf H_m$ is an orthogonally normalized Hadamard matrix, each $\mathbf D_j$ is diagonal with independent random signs in $\{-1,+1\}$, and $\gamma$ sets the projection scale.
The transform dimension $m\geq N+d$ is a power of two, and the corresponding $N\times(N+d)$ submatrix gives $\mathbf S$.
The fast Hadamard transform evaluates this projection in $O(m\log m)$ operations without storing a dense matrix.
The nonlinear state update still processes the sequence one time step at a time.

\subsection{Parallel Echo State Network (ParalESN)}
\label{app:paralesn}
ParalESN separates linear recurrence from nonlinear mixing~\citep{pinna2026paralesn}.
For one layer with input $\mathbf u_t$, the complex-valued state follows
\begin{equation}
\begin{aligned}
\mathbf h_t&=\overline{\mathbf\Lambda}\mathbf h_{t-1}
+\alpha(\mathbf W_{\mathrm{in}}\mathbf u_t+\mathbf b),\\
\overline{\mathbf\Lambda}&=(1-\alpha)\mathbf I+\alpha\mathbf\Lambda.
\end{aligned}
\label{eq:app-paralesn}
\end{equation}
Here, $\mathbf\Lambda$ is a fixed complex diagonal matrix, $\mathbf W_{\mathrm{in}}$ is the input matrix, $\mathbf b$ is a bias, $\mathbf I$ is the identity matrix, and $\alpha\in(0,1]$ is the leak rate.
The diagonal entries of the effective transition matrix $\overline{\mathbf\Lambda}$ control decay and rotation of the state coordinates.
The mixer then produces real-valued features:
\begin{equation}
\mathbf q_t=\tanh\!\left(\operatorname{Re}(\mathbf C\mathbf h_t+\mathbf c)\right),
\label{eq:app-paralesn-mixer}
\end{equation}
where $\mathbf C$ represents a fixed convolution across state coordinates, $\mathbf c$ is its bias, and $\operatorname{Re}$ extracts the real part.
These features feed the next block and the trainable readout.
The mixer acts at each time step, outside the temporal recurrence in Eq.~\ref{eq:app-paralesn}.
The recurrence therefore remains affine, so its updates compose associatively and can be evaluated by a parallel scan over a given input sequence.

\begin{figure*}[t]
\centering
\appendixArchitectureTable
\par\bigskip
\includegraphics[width=\textwidth]{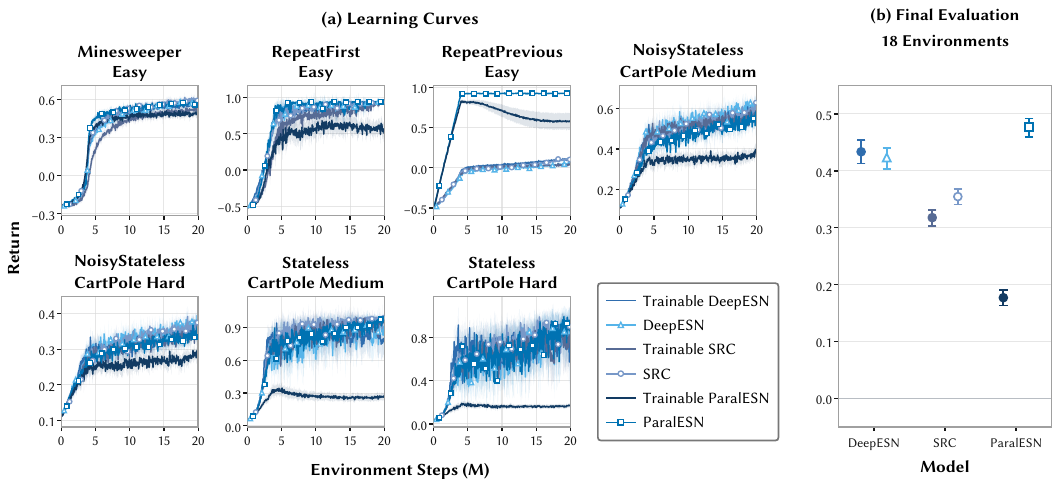}
\caption{POPGym results for DeepESN, SRC, and ParalESN and their trainable variants. (a) Learning curves on the selected seven tasks  over ten seeds. (b) Final evaluation returns across all $18$ POPGym tasks over ten seeds and ten evaluation episodes per seed}
\label{fig:app-esn}
\Description{The left panel contains seven training-return plots and a shared legend for DeepESN, SRC, and ParalESN and their trainable variants. The right panel compares final evaluation returns averaged across eighteen POPGym settings, with bootstrap confidence intervals. Filled markers show trainable models, and hollow markers show standard models.}
\end{figure*}

\section{Reservoir Training on POPGym}
\label{app:esn}
We compare DeepESN, SRC, and ParalESN with their trainable variants on the 18 POPGym tasks.
Fig.~\ref{fig:app-esn}(a) shows the learning curves, and Fig.~\ref{fig:app-esn}(b) summarizes final evaluation returns across all tasks.
\section{Detailed Evaluation Results}
\label{app:evaluation}
We report detailed final evaluation returns for POPGym, POPGym Arcade, and MuJoCo Playground.
For each seed, we evaluate the trained agent over 10 episodes and compute its mean return.
Each heatmap entry shows the mean and sample standard deviation across seeds.
We use 10 seeds for POPGym and MuJoCo Playground, and 5 for POPGym Arcade.

\subsection{POPGym}
\begin{figure*}[p]
\centering
\appendixEncoderTable
\par\bigskip
\includegraphics[width=\textwidth]{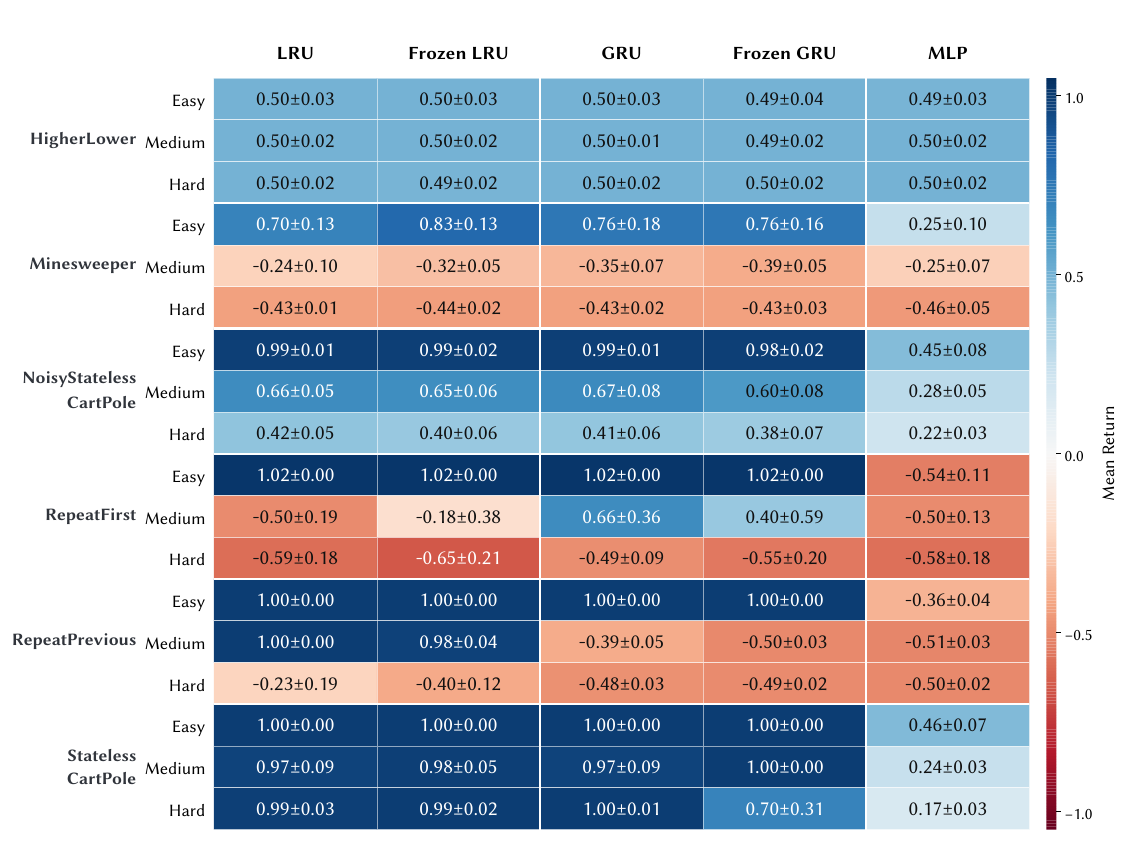}
\caption{Detailed final evaluation returns for LRU, GRU, Frozen GRU , Frozen LRU and MLP on POPGym. Each entry shows the mean $\pm$ sample standard deviation over 10 seeds, with 10 evaluation episodes per seed}
\label{fig:app-eval-recurrent}
\Description{A heatmap reports final evaluation returns for MLP and variants of GRU and LRU across eighteen POPGym settings. Each cell contains a mean and standard deviation. Columns distinguish trainable models from variants with frozen recurrent cores.}
\end{figure*}

\begin{figure*}[p]
\centering
\appendixMemoryChainPopgymTables
\par\bigskip
\includegraphics[width=\textwidth]{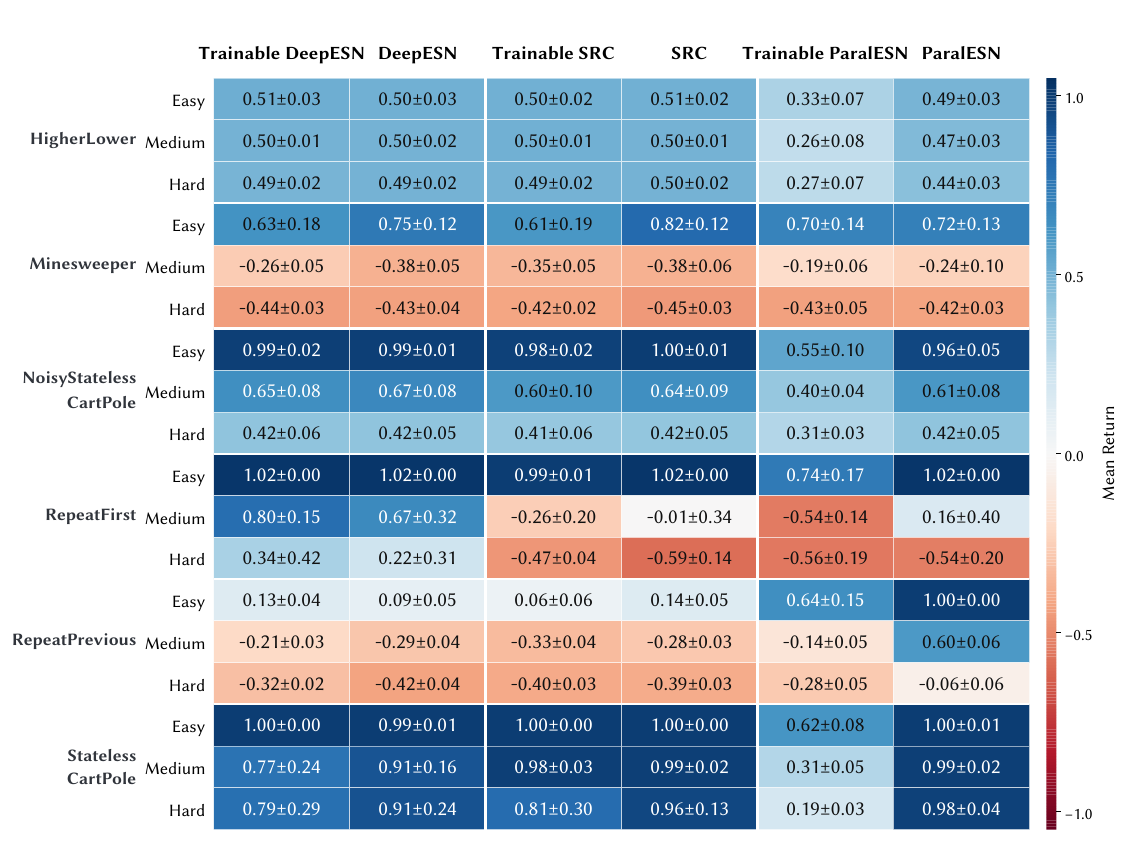}
\caption{Detailed final evaluation returns for DeepESN, SRC, ParalESN, and their trainable variants on POPGym. Each entry shows the mean $\pm$ sample standard deviation over 10 seeds, with 10 evaluation episodes per seed.}
\label{fig:app-eval-reservoir}
\Description{A heatmap compares final evaluation returns for DeepESN, SRC, and ParalESN and their trainable variants across eighteen POPGym settings. Each cell contains a mean and standard deviation.}
\end{figure*}

The POPGym heatmaps report evaluation results for $18$ tasks and $11$ model configurations.
Figs.~\ref{fig:app-eval-recurrent} and~\ref{fig:app-eval-reservoir} report the final evaluation returns for the recurrent baselines and reservoir models, respectively.
Both heatmaps use a shared color scale centered at zero.
Frozen GRU and Frozen LRU retain trainable mixers and outer projections.
We exclude Autoencode, Concentration, CountRecall, and MultiArmedBandit because of implementation bugs in the vector environments.
ParalESN achieves a higher mean final evaluation return than its trainable counterpart on six of the seven settings selected for the learning curves (Fig.~\ref{fig:app-eval-reservoir}).

\subsection{POPGym Arcade and MuJoCo Playground}
\begin{figure*}[p]
\centering
\appendixArcadePlaygroundTables
\par\bigskip
\includegraphics[width=\textwidth]{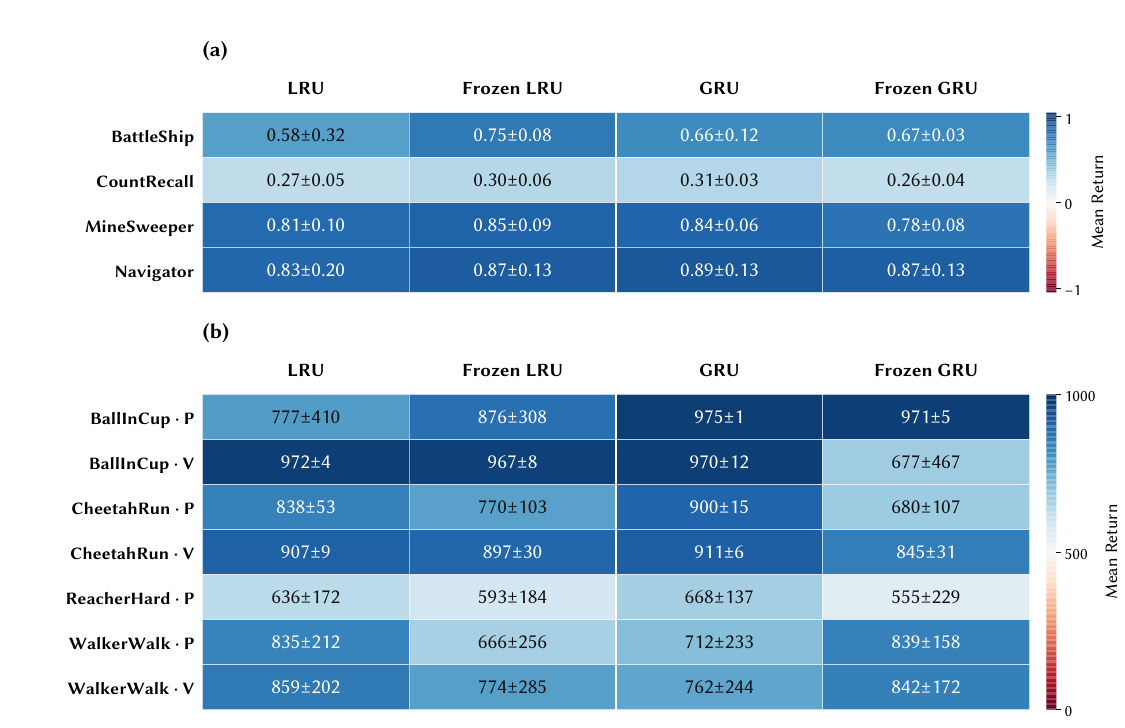}
\caption{Detailed final evaluation returns for GRU, LRU, and their frozen variants on POPGym Arcade (top) and MuJoCo Playground (bottom). Each entry shows the mean $\pm$ sample standard deviation over 5 and 10 seeds, respectively, with 10 evaluation episodes per seed.}
\label{fig:app-eval-other}
\Description{The top heatmap reports final evaluation returns for four GRU and LRU configurations on four POPGym Arcade tasks. The bottom heatmap reports the same configurations on seven MuJoCo Playground settings. Each cell shows a mean and standard deviation, and the panels use separate color scales.}
\end{figure*}

We report the final evaluation returns of GRU, LRU, and their frozen variants on four POPGym Arcade tasks and seven MuJoCo Playground settings in Fig.~\ref{fig:app-eval-other}.
We train the Arcade agents with PQN and the Playground agents with PPO.
POPGym Arcade uses a separate implementation from vector POPGym, so we retain its CountRecall task.

\section{Model and Training Parameters}
\label{app:parameters}
Tables~\ref{tab:appendix-architecture} and~\ref{tab:appendix-arcade-encoder} describe the recurrent agents and the CNN encoder used for POPGym Arcade.
Here, $d_o$ is the vector observation dimension, $d_a$ is the number of discrete actions or the continuous action dimension, and $d_{\mathrm{in}}$ is the input feature dimension.
We use $L=1$ recurrent block for MemoryChain and $L=2$ for the other benchmarks.
Each readout has $K=1$ hidden block for MemoryChain, $K=2$ for POPGym and POPGym Arcade, and $K=3$ for MuJoCo Playground.
PPO uses separate actor and critic readouts with a shared recurrent trunk.
All LeakyReLU activations use a negative slope of $0.01$.
The CNN uses no padding, and its output feature-map dimensions are listed as channels $\times$ height $\times$ width.

We use Adam with linear learning-rate decay.
Tables~\ref{tab:appendix-training-memorychain}--\ref{tab:appendix-training-playground} list the main training hyperparameters.
For MuJoCo Playground, we report the range of initial learning rates used across models and tasks.

\end{document}